\documentclass[10pt,twocolumn,letterpaper]{article}

\usepackage[pagenumbers]{cvpr} %
\usepackage[accsupp]{axessibility}

\usepackage{graphicx}
\usepackage{amsmath}
\usepackage{amssymb}
\usepackage{booktabs}
\usepackage{tabularray}
\usepackage{multicol}
\usepackage{multirow}
\usepackage{caption}
\usepackage{algorithm2e}
\usepackage{algpseudocode}
\usepackage{comment}
\usepackage[symbol]{footmisc}
\RestyleAlgo{ruled}
\usepackage[dvipsnames]{xcolor}
\usepackage{longtable}
\usepackage{tocloft}
\UseTblrLibrary{booktabs}

\usepackage[dvipsnames]{xcolor}

\newcommand{\Sref}[1]{Sec.~\ref{#1}}
\newcommand{\Eref}[1]{Eq.~(\ref{#1})}
\newcommand{\Fref}[1]{Fig.~\ref{#1}}
\newcommand{\Tref}[1]{Table~\ref{#1}}
\newcommand{\R}[0]{\mathbb{R}}
\newcommand{\x}[0]{\times}

\newcommand\lft{\mathopen{}\left}
\newcommand\rgt{\aftergroup\mathclose\aftergroup{\aftergroup}\right}
\newcommand{\fst}[1]{\textbf{#1}}
\newcommand{\snd}[1]{\underline{#1}}
\newcommand{\std}[1]{\tiny{(#1)}}
\newcommand\mypara[1]{\vspace{1mm}\noindent\textbf{#1}}

\definecolor{gainsboro}{rgb}{0.86, 0.86, 0.86}
\definecolor{myblue}{rgb}{0,0.6902,0.94118}

\DefTblrTemplate{caption-tag}{default}{Table \thetable}
\DefTblrTemplate{caption-sep}{default}{.\enskip}

\NewTblrTheme{cvpr}{
 \SetTblrStyle{head}{font=\small}
 \SetTblrStyle{caption}{hang=0pt,indent=0pt}
 \SetTblrStyle{conthead}{hang=0pt,indent=0pt}
 \SetTblrStyle{firsthead}{hang=0pt,indent=0pt}
 \SetTblrStyle{middlehead}{hang=0pt,indent=0pt}
 \SetTblrStyle{lasthead}{hang=0pt,indent=0pt}

}

\definecolor{cvprblue}{rgb}{0.21,0.49,0.74}
\usepackage[pagebackref,breaklinks,colorlinks,citecolor=cvprblue]{hyperref}

\def\paperID{10140} %
\def\confName{CVPR}
\def\confYear{2024}

\title{MaGGIe: Masked Guided Gradual Human Instance Matting}

\author{
Chuong Huynh$^{1}$\footnotemark \quad Seoung Wug Oh$^{2}$ \quad Abhinav Shrivastava$^{1}$\quad Joon-Young Lee$^{2}$  \\
$^1$University of Maryland, College Park \quad  $^2$Adobe Research \\
 $^1${\tt\small \{chuonghm,abhinav\}@cs.umd.edu}\quad $^2${\tt\small \{seoh,jolee\}@adobe.com}
}

\begin{document}
\maketitle
\setcounter{tocdepth}{-1}
\begin{abstract}

Human matting is a foundation task in image and video processing where human foreground pixels are extracted from the input. Prior works either improve the accuracy by additional guidance or improve the temporal consistency of a single instance across frames. We propose a new framework \textbf{MaGGIe}, \textbf{Ma}sked \textbf{G}uided \textbf{G}radual Human \textbf{I}nstanc\textbf{e} Matting, which predicts alpha mattes progressively for each human instances while maintaining the computational cost, precision, and consistency. Our method leverages modern architectures, including transformer attention and sparse convolution, to output all instance mattes simultaneously without exploding memory and latency. Although keeping constant inference costs in the multiple-instance scenario, our framework achieves robust and versatile performance on our proposed synthesized benchmarks. With the higher quality image and video matting benchmarks, the novel multi-instance synthesis approach from publicly available sources is introduced to increase the generalization of models in real-world scenarios. Our code and datasets are available at \url{https://maggie-matt.github.io}.

\end{abstract}
\footnotetext[1]{This work was done during Chuong Huynh's internship at Adobe}
   
\section{Introduction}

\begin{figure}[t]
    \centering
    \includegraphics[width=\columnwidth]{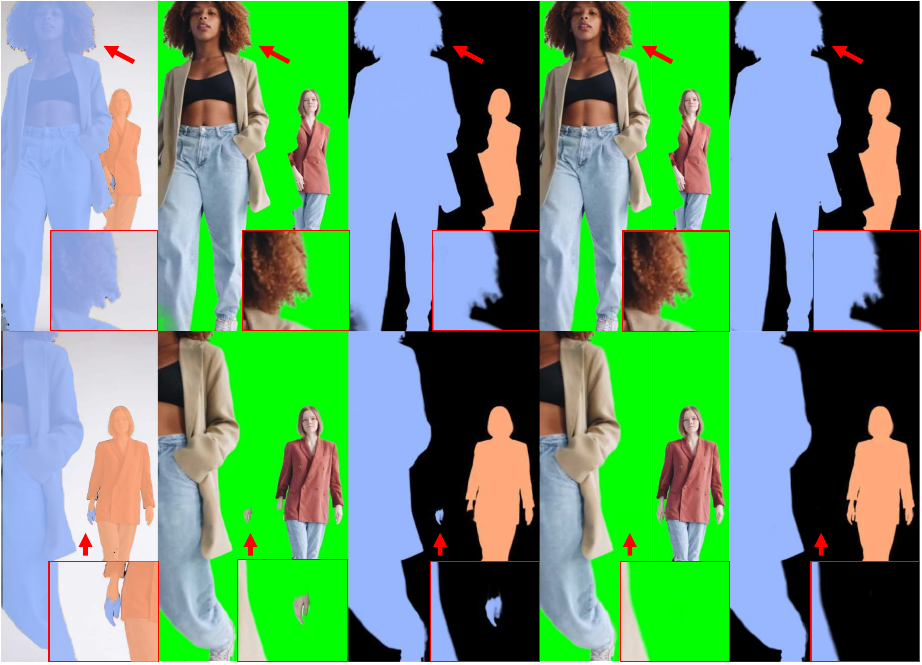}
    \begin{tblr}{width=\columnwidth,colsep=1pt,rowsep=3pt,colspec={@{}X[1.0,c]X[2.5,c]X[2.5,c]@{}}}
        Guidance masks 
        & \raisebox{2mm}{$\underbrace{\hspace{3.3cm}}_{\colorbox{white}{~~InstMatt~\cite{sun2022instmatt}~~}}$}
         & \raisebox{2mm}{$\underbrace{\hspace{3.3cm}}_{\colorbox{white}{~~MaGGIe (ours)~~}}$}
    \end{tblr}
    \vspace{-1em}
    \caption{\textbf{Our MaGGIe delivers precise and temporally consistent alpha mattes.} It adeptly preserves intricate details and demonstrates robustness against noise in instance guidance masks by effectively utilizing information from adjacent frames. Red arrows highlight the areas of detailed zoom-in. (Optimally viewed in color and digital zoom in).}
    \vspace{-1.0em}
    \label{fig:teaser}
\end{figure}

In image matting, a trivial solution is to predict the pixel transparency - alpha matte $\alpha \in [0,1]$ for precise background removal. Considering a saliency image $I$ with two main components, foreground $F$ and background $B$, the image $I$ is expressed as $I=\alpha F + (1-\alpha) B$.
Because of the ambiguity in detecting the foreground region, for example, whether a person's belongings are a part of the human foreground or not, many methods~\cite{hou2019context,li2020gca,lu2019indexnet,forte2020fba} leverage additional guidance, typically trimaps, defining foreground, background, and unknown or transition regions. However, creating trimaps, especially for videos, is resource-intensive. Alternative binary masks~\cite{yu2021mgm,park2023mgmwild} are simpler to obtain by human drawings or off-the-shelf segmentation models while offering greater flexibility without hardly constraint output values of regions as trimaps.
Our work focuses but is not limited to human matting because of the higher number of available academic datasets and user demand in many applications~\cite{cutoutsios16,repositionpixel8,sengupta2020background,adobepremiere,hebborn2017occlusion} compared to other objects.

When working with video input, the problem of creating trimap guidance is often resolved by guidance propagation~\cite{seong2022otvm,huang2023ftpvm} where the main idea coming from video object segmentation~\cite{oh2019stm,cheng2022xmem}. However, the performance of trimap propagation degrades when video length grows. The failed trimap predictions, which miss some natures like the alignment between foreground-unknown-background regions, lead to incorrect alpha mattes. We observe that using binary masks for each frame gives more robust results. However, the consistency between the frame's output is still important for any video matting approach. For example, holes appearing in a random frame because of wrong guidance should be corrected by consecutive frames. Many works~\cite{lin2022rvm,seong2022otvm,huang2023ftpvm,lin2023adam,wang2023hstg} constrain the temporal consistency at feature maps between frames. Since the alpha matte values are very sensitive, feature-level aggregation is not an absolute guarantee of the problem. Some methods~\cite{ke2022vmt,sun2023sparsemat} in video segmentation and matting compute the incoherent regions to update values across frames. We propose a temporal consistency module that works in both feature and output spaces to produce consistent alpha mattes. 

Instance matting~\cite{sun2022instmatt} is an extension of the matting problem where there exists multiple $\alpha_i, i \in {0..N}$, and each belongs to one foreground instance. This problem creates another constraint for each spatial location $(x,y)$ value such that $\sum_i \alpha_i (x, y) = 1$. The main prior work InstMatt~\cite{sun2022instmatt} handles the multi-instance images by predicting each alpha matte separately from binary guided masks before the instance refinement at the end. Although this approach produces impressive results in both synthesized and natural image benchmarks, the efficiency and accuracy of this model are unexplored in video processing. The separated prediction for each instance yields inefficiency in the architecture, which makes it costly to adapt to video input. Another concurrent work~\cite{li2024vim} with ours extends the InstMatt to process video input, but the complexity and efficiency of the network are unexplored. \Fref{fig:teaser} illustrates the comparison between our MaGGIe and InstMatt when working with video. Our work improves not only the accuracy but also the consistency between frames when errors occur in guidance.

Besides the temporal consistency, when extending the instance matting to videos containing a large number of frames and instances, the careful network design to prevent the explosion in the computational cost is also a key challenge. In this work, we propose several adjustments to the popular mask-guided progressive refinement architecture~\cite{yu2021mgm}. Firstly, by using the mask guidance embedding inspired by AOT~\cite{yang2021aot}, the input size reduces to a constant number of channels. Secondly, with the advancement of transformer attention in various vision tasks~\cite{phung2024attenref,pham2024cora,pham2022improving}, we inherit the query-based instance segmentation~\cite{kirillov2023sam,huynh2023simpson,cheng2022mask2former} to predict instance mattes in one forward pass instead of separated estimation. It also replaces the complex refinement in previous work with the interaction between instances by attention mechanism. To save the high cost of transformer attention, we only perform multi-instance prediction at the coarse level and adapt the progressive refinement at multiple scales~\cite{yu2021mgm,huynh2021magnet}. However, using full convolution for the refinement as previous works are inefficient as less than 10\% of values are updated at each scale, which is also mentioned in~\cite{sun2023sparsemat}. The replacement of sparse convolution~\cite{liu2015sparse} saves the inference cost significantly, keeping the constant complexity of the algorithm since only interested locations are refined. Nevertheless, the lack of information at a larger scale when using sparse convolution can cause a dominance problem, which leads to the higher-scale prediction copying the lower outputs without adding fine-grained details. We propose an instance guidance method to help the coarser prediction guide but not contribute to the finer alpha matte.

In addition to the framework design, we propose a new training video dataset and benchmarks for instance-awareness matting. Besides the new large-scale high-quality synthesized image instance matting, an extension of the current instance image matting benchmark adds more robustness with different guidance quality. For video input, our synthesized training and benchmark are constructed from various public instance-agnostic datasets with three levels of difficulty. 

In summary, our contributions include:
\begin{itemize}
\item A highly efficient instance matting framework with mask guidance that has all instances interacting and processed in a single forward pass.
\item A novel approach that considers feature-matte levels to maintain matte temporal consistency in videos.
\item Diverse training datasets and robust benchmarks for image and video instance matting that bridge the gap between synthesized and natural cases.
\end{itemize}

\section{Related Works}
There are many ways to categorize matting methods, here we revise previous works based on their primary input types. The brief comparison of others and our MaGGIe is shown in \Tref{tab:summary}. 

\mypara{Image Matting.} Traditional matting methods~\cite{levin2007closed,lee2011nonlocal,berman2000method} rely on color sampling to estimate foreground and background, often resulting in noisy outcomes due to limited high-level object features. Advanced deep learning-based methods~\cite{cho2016natural,shen2016deep,xu2017dim,sun2021sim,lu2019indexnet,li2020gca,forte2020fba} have significantly improved results by integrating image and trimap inputs or focusing on high-level and detailed feature learning. However, these methods often struggle with trimap inaccuracies and assume single-object scenarios. Recent approaches~\cite{ke2022modnet,chen2022ppmatting,chen2022sghm} require only image inputs but face challenges with multiple salient objects. MGM~\cite{yu2021mgm} and its extension MGM-in-the-wild~\cite{park2023mgmwild} introduce binary mask-based matting, addressing multi-salient object issues and reducing trimap dependency. InstMatt~\cite{sun2022instmatt} further customizes this approach for multi-instance scenarios with a complex refinement algorithm. Our work extends these developments, focusing on efficient, end-to-end instance matting with binary mask guidance.
Image matting also benefits from diverse datasets~\cite{xu2017dim,li2021aim500,li2021ppm,li2022am2k,ke2022modnet,lin2021bgmv2,sun2023sparsemat}, supplemented by background augmentation from sources like BG20K~\cite{li2022am2k} or COCO~\cite{lin2014coco}. Our work also leverages currently available datasets to concretize a robust benchmark for human-masked guided instance matting.

\mypara{Video Matting.} Temporal consistency is a key challenge in video matting. Trimap-propagation methods~\cite{sun2021dvm,seong2022otvm,huang2023ftpvm} and background knowledge-based approaches like BGMv2~\cite{lin2021bgmv2} aim to reduce trimap dependency. Recent techniques~\cite{lin2022rvm,lin2023adam,wang2023hstg,zhang2021tcvom,li2022vmformer} incorporate Conv-GRU, attention memory matching, or transformer-based architectures for temporal feature aggregation. SparseMat~\cite{sun2023sparsemat} uniquely focuses on fusing outputs for consistency. Our approach builds on these foundations, combining feature and output fusion for enhanced temporal consistency in alpha maps.
There is a lack of video matting datasets due to the difficulty in data collecting. VideoMatte240K~\cite{lin2021bgmv2} and VM108~\cite{zhang2021tcvom} focus on composited videos, while CRGNN~\cite{wang2021crgnn} is the only offering natural videos for human matting. To address the gap in instance-aware video matting datasets, we propose adapting existing public datasets for training and evaluation, particularly for human subjects.

\begin{table}[t!]
    \centering
    \scriptsize
    \caption{\textbf{Comparing MaGGIe with previous works in image and video matting.} Our work is the first instance-aware framework producing alpha matte from a binary mask with both feature and output temporal consistency in constant processing time.}
    \begin{tblr}{width=\columnwidth,colsep=2.5pt,colspec={@{}l|cccccc}}
    \toprule
         \SetCell[r=2]{l}{Method}
         & \SetCell[r=2]{c}{Avenue}
            & \SetCell[r=2]{c}{Guidance}
                & \SetCell[r=2]{c}{Instance\\-awareness}
                    & \SetCell[c=2]{c}{Temp. aggre.} & 
                        & \SetCell[r=2]{c}{Time\\ complexity} \\
    \hline
    & & & & Feat. & Matte. \\
    \hline
    MGM~\cite{yu2021mgm,park2023mgmwild} & CVPR21+23 & Mask & & & & $O(n)$ \\
    InstMatt~\cite{sun2022instmatt} & CVPR22 & Mask & \checkmark & & & $O(n)$ \\
    \hline
    TCVOM~\cite{zhang2021tcvom} & MM21 & - & - & \checkmark & & - \\
    OTVM~\cite{seong2022otvm} & ECCV22 & 1st trimap & & \checkmark & & $O(n)$ \\
    FTP-VM~\cite{huang2023ftpvm} & CVPR23 & 1st trimap & & \checkmark & & $O(n)$ \\
    SparseMatt~\cite{sun2023sparsemat} & CVPR23 & No & & & \checkmark & $O(n)$ \\
    \hline
    MaGGIe & - & Mask & \checkmark & \checkmark & \checkmark & $\approx O(1)$ \\
    \bottomrule
    \end{tblr}
    \label{tab:summary}
    \vspace{-1.5em}
\end{table}

\section{MaGGIe}

\begin{figure*}
    \centering
    \vspace{-2em}
    \includegraphics[width=0.98\textwidth]{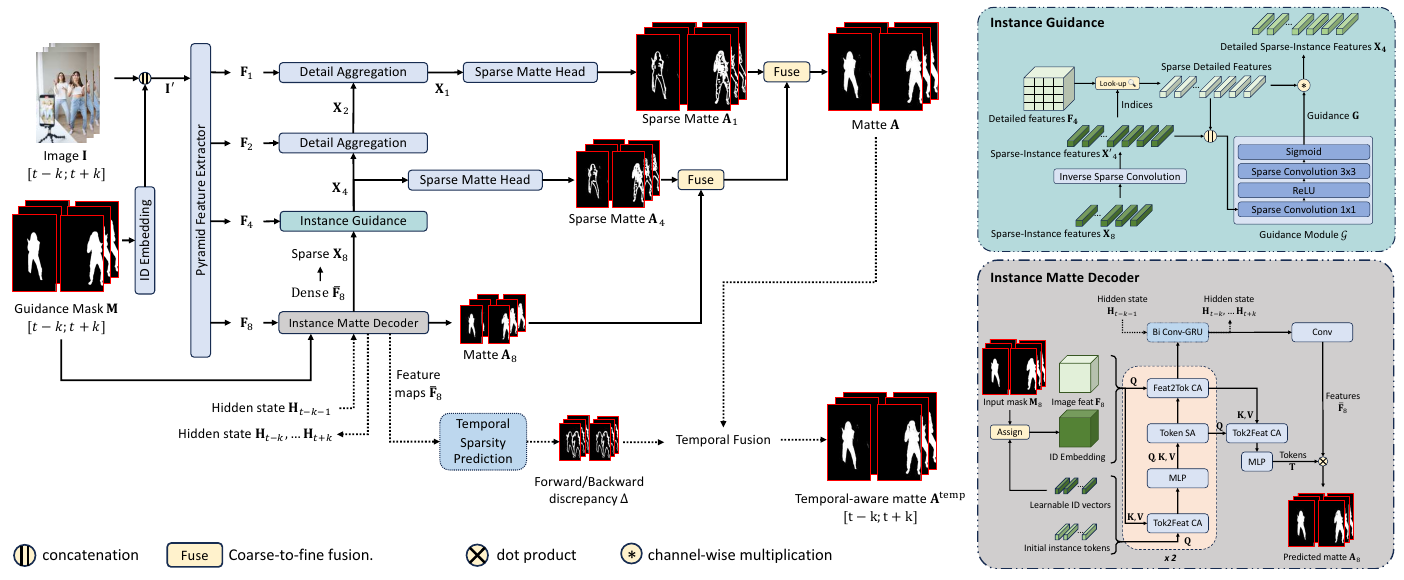}
    \vspace{-0.5em}
    \caption{\textbf{Overall pipeline of MaGGIe}. This framework processes frame sequences $\mathbf{I}$ and instance masks $\mathbf{M}$ to generate per-instance alpha mattes $\mathbf{A}'$ for each frame. It employs progressive refinement and sparse convolutions for accurate mattes in multi-instance scenarios, optimizing computational efficiency. The subfigures on the right illustrate the Instance Matte Decoder and the Instance Guidance, where we use mask guidance to predict coarse instance mattes and guide detail refinement by deep features, respectively. (Optimal in color and zoomed view).}
    \label{fig:method_overall}
    \vspace{-1.0em}
\end{figure*}

We introduce our efficient instance matting framework guided by instance binary masks, structured into two parts. The first~\Sref{sec:method_framework} details our novel architecture to maintain accuracy and efficiency. The second~\Sref{sec:temp_cons} describes our approach for ensuring temporal consistency across frames in video processing.

\subsection{Efficient Masked Guided Instance Matting}
\label{sec:method_framework}
Our framework, depicted in \Fref{fig:method_overall}, processes images or video frames $\mathbf{I}\in [0,255]^{T\x 3 \x H \x W}$ with corresponding binary instance guidance masks $\mathbf{M}\in \{0,1\}^{T \x N \x H \x W}$, and then predicts alpha mattes $\mathbf{A} \in [0,1]^{T \x N \x H \x W}$ for each instance per frame. Here, $T, N, H, W$ represent the number of frames, instances, and input resolution, respectively. Each spatial-temporal location $(x,y,t)$ in $\mathbf{M}$ is a one-hot vector $\{0,1\}^N$ highlighting the instance it belongs to. The pipeline comprises five stages: (1) Input construction; (2) Image features extraction; (3) Coarse instance alpha mattes prediction; (4) Progressive detail refinement; and (5) Coarse-to-fine fusion.

\mypara{Input Construction.} The input $\mathbf{I}^\prime \in \R^{T\x (3 + C_e) \x H \x W}$ to our model is the concatenation of input image $\mathbf{I}^\prime$ and guidance embedding $\mathbf{E} \in \R^{T\x C_e \x H \x W}$ constructed from $\mathbf{M}$ by ID Embedding layer~\cite{yang2021aot}. More details about transforming $\mathbf{M}$ to $\mathbf{E}$ are in the supplementary material.

\mypara{Image Features Extraction.} We extract features map $\mathbf{F}_s \in \R^{T \x C_s \x H/s \x W/s}$ from $\mathbf{I}^\prime$ by feature-pyramid networks. As shown in the left part of~\Fref{fig:method_overall}, there are four scales $s={1,2,4,8}$ for our coarse-to-fine matting pipeline.

\mypara{Coarse instance alpha mattes prediction.} Our MaGGIe adopts transformer-style attention to predict instance mattes at the coarsest features $\mathbf{F}_8$. We revisit the scaled dot-product attention mechanism in Transformers~\cite{vaswani2017attention}. Given queries $\mathbf{Q} \in \R^{L \x C}$, keys $\mathbf{K} \in \R^{S \x C}$, and values $\mathbf{V} \in \R^{S \x C}$, the scaled dot-product attention is defined as:
\begin{equation}
    \text{Attention}(\mathbf{Q}, \mathbf{K}, \mathbf{V}) = \text{softmax}\left(\frac{\mathbf{Q}\mathbf{K}^\top}{\sqrt{C}}\right)\mathbf{V}.
\end{equation}
In cross-attention (CA), $\mathbf{Q}$ and $(\mathbf{K}, \mathbf{V})$ originate from different sources, whereas in self-attention (SA), they share similar information.

In our Instance Matte Decoder, the organization of CA and SA blocks inspired by SAM~\cite{kirillov2023sam} is depicted in the bottom right of~\Fref{fig:method_overall}. The downscaled guidance masks $\mathbf{M}_8$ also participate as the additional embedding for image features in attention procedures. The coarse alpha matte  $\mathbf{A}_8$ is computed as the dot product between instance tokens $\mathbf{T} = \{\mathbf{T}_i| 1 \leq i \leq N\} \in \R^{N \x C_8}$ and enriched feature map $\bar{\mathbf{F}}_8$ with a sigmoid activation applied. Those components are used in the following steps of matte detail refinement.

\mypara{Progressive Detail Refinement.} From the coarse instance alpha matte, we leverage the Progressive Refinement~\cite{yu2021mgm} to improve the details at uncertain locations $\mathbf{U} = \{u_p=(x,y,t, i)|0<\mathbf{A}_8(u_p)<1\} \in \mathbb{N}^{P\x 4}$ with some highly efficient modifications. It is mandatory to transform enriched dense features $\bar{\mathbf{F}}_8$ to instance-specific features $\mathbf{X}_8$ for the instance-wise refinement. However, to save memory and computational costs, only transformed features at uncertainty $\mathbf{U}$ are computed as:
\begin{equation}
\mathbf{X}_8(x, y, t, i) = \text{MLP}(\mathbf{\bar{F}}_8(x, y, t) \times \mathbf{T}_i).
\label{eq:dense2sparse}
\end{equation}

To combine the coarser instance-specific sparse features $\mathbf{X}_8$ with the finer image features $\mathbf{F}_4$, we propose the Instance Guidance (IG) module. As described in the top right of~\Fref{fig:method_overall}, this module firstly increases the spatial scale of $\mathbf{X}_8$ to have $\mathbf{X}^\prime_4$ by an inverse sparse convolution. For each entry $p$, we compute a guidance score $\mathbf{G} \in [0,1]^{C_4}$, which is then channel-wise multiplied with $\mathbf{F}_4$ to produce detailed sparse instance-specific features $\mathbf{X}_4$:
\begin{equation}
\mathbf{X}_4(p) = \mathcal{G}\lft(\lft\{\mathbf{X}'_4(p);\mathbf{F}_4(p)\rgt\}\rgt) \times \mathbf{F}_4(p),
\end{equation}
where $\{;\}$ denotes concatenation along the feature dimension, and $\mathcal{G}$ is a series of sparse convolutions with sigmoid activation.

The sparse features $\mathbf{X}_4$ is then aggregated with other dense features $\mathbf{F}_2, \mathbf{F}_1$ respectively at corresponding indices to have $\mathbf{X}_2, \mathbf{X}_1$. At each scale, we predict alpha matte $\mathbf{A}_4,\mathbf{A}_1$ with gradual detail improvement. You can find more aggregation and sparse matting head details in the supplementary material. 

\mypara{Coarse-to-fine fusion.} This stage is to combine alpha mattes of different scales in a progressive way (PRM): $\mathbf{A}_8 \rightarrow \mathbf{A}_4 \rightarrow \mathbf{A}_1$ to obtain $\mathbf{A}$. At each step, only values at uncertain locations and belonging to unknown masks are refined.

\mypara{Training Losses.} In addition to standard losses ($\mathcal{L}_1$ for reconstruction, Laplacian $\mathcal{L}_{\text{lap}}$ for detail, Gradient $\mathcal{L}_{\text{grad}}$ for smoothness), we supervise the affinity score matrix Aff between instance tokens $\mathbf{T}$ (as $\mathbf{Q}$) and image feature maps $\mathbf{F}$ (as $\mathbf{K,V}$) by the attention loss $\mathcal{L}_{\text{att}}$. Additionally, our network's progressive refinement process necessitates accurate coarse-level predictions to determine $\mathbf{U}$ accurately. We assign customized weights $\mathbf{W}_8$ for losses at scale $s=8$ to prioritize uncertain locations. More details about $\mathcal{L}_{\text{att}}$ and $\mathbf{W}_8$ is in the supplementary material.

\subsection{Feature-Matte Temporal Consistency}
\label{sec:temp_cons}
We propose to enhance temporal consistency at both feature and alpha matte levels.

\mypara{Feature Temporal Consistency.} Utilizing Conv-GRU~\cite{ballas2015convgru} for video inputs, we ensure bidirectional consistency among feature maps of adjacent frames. With a temporal window size $k$, bidirectional Conv-GRU processes frames $\{t-k, ... t+k\}$, as shown in \Fref{fig:method_overall}. For simplicity, we set $k=1$ with an overlap of 2 frames. The initial hidden state $\mathbf{H}_0$ is zeroed, and $\mathbf{H}_{t-k-1}$ from the previous window aids the current one. This module fuses the feature map at time $t$ with two consecutive frames, averaging forward and backward aggregations. The resultant temporal features are used to predict the coarse alpha matte $\mathbf{A}_8$.

\mypara{Alpha Matte Temporal Consistency.} We propose fusing frame mattes by predicting their temporal sparsity. Unlike the previous method~\cite{sun2023sparsemat} using image processing kernels, we leverage deep features for this prediction. A shallow convolutional network with a sigmoid activation processes stacked feature maps $\mathbf{\bar{F}}_8$ at $t-1$ and $t$, outputting alpha matte discrepancy between two frames $\Delta(t) \in \{0,1\}^{H \x W}$. For each frame $t$, with $\Delta(t)$ and $\Delta(t+1)$, we compute the forward propagation $\mathbf{A}^f$ and backward propagation $\mathbf{A}^b$ to reject the propagation at misalignment regions and obtain temporal aware output $\mathbf{A}^{\text{temp}}$. The supplementary material provides more details about the implementation.

\mypara{Training Losses.} Besides the dtSSD loss for temporal consistency, we introduce an L1 loss for the alpha matte discrepancy. The loss compares predicted $\Delta (t)$ with the ground truth $\Delta^{gt}(t) = \max_i \left(|\mathbf{A}^{gt}(t-1, i) -\mathbf{A}^{gt}(t, i)| > \beta\right)$, where $\beta=0.001$ to simplify the problem to binary pixel classification.

\section{Instance Matting Datasets}

\begin{figure}[t]
    \centering
    \vspace{-1.5em}
    \includegraphics[width=\columnwidth]{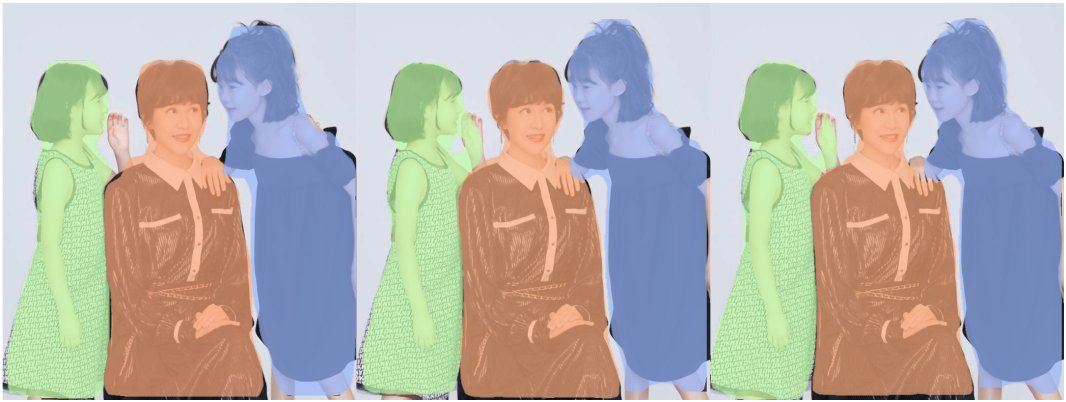}
    \caption{\textbf{Variations of Masks for the Same Image in M-HIM2K Dataset.} Masks generated using R50-C4-3x, R50-FPN-3x, R101-FPN-400e MaskRCNN models trained on COCO. (Optimal in color).}
    \label{fig:data_mask}
    \vspace{-1.5em}
\end{figure}

\begin{table}[b]
    \centering
    \vspace{-1em}
    \caption{\textbf{Details of Video Instance Matting Training and Testing Sets.} V-HIM2K5 for training and V-HIM60 for model evaluation. Each video contains 30 frames.}
    \footnotesize
    \begin{tblr}{width=\columnwidth,colsep=3pt,colspec={@{}l|ccc|ccc|ccc@{}}}
    \toprule
         \SetCell[r=2]{c}{Name} 
         & \SetCell[c=3]{c}{Sources} & & 
         & \SetCell[c=3]{c}{\# videos} & &
         & \SetCell[c=3]{c}{\# instance/video} \\
         \hline
    & \cite{zhang2021tcvom} & \cite{lin2021bgmv2} & \cite{wang2021crgnn} & Easy & Med. & Hard & Easy & Med. & Hard \\
    \hline
    V-HIM2K5 & 33 & 410 & 0 & 500 & 1,294 & 667 & 2.67 & 2.65 & 3.21 \\
    V-HIM60 & 3 & 8 & 18 & 20 & 20 & 20 & 2.35 & 2.15 & 2.70 \\
    \bottomrule
    \end{tblr}
    \label{tab:data_vid_info}
\end{table}

This section outlines the datasets used in our experiments. With the lack of public datasets for the instance matting task, we synthesized training data from existing public instance-agnostic sources. Our evaluation combines synthetic and natural sets to assess the model's robustness and generalization.

\subsection{Image Instance Matting}

We derived the \textbf{Image Human Instance Matting 50K (I-HIM50K)} training dataset from HHM50K~\cite{sun2023sparsemat}, featuring multiple human subjects. This dataset includes 49,737 synthesized images with 2-5 instances each, created by compositing human foregrounds with random backgrounds and modifying alpha mattes for guidance binary masks. For benchmarking, we used HIM2K~\cite{sun2022instmatt} and created the \textbf{Mask HIM2K (M-HIM2K)} set to test robustness against varying mask qualities from available instance segmentation models (as shown in~\Fref{fig:data_mask}). Details on the generation process are available in the supplementary material.

\subsection{Video Instance Matting}

Our video instance matte dataset, synthesized from VM108~\cite{zhang2021tcvom}, VideoMatte240K~\cite{lin2021bgmv2}, and CRGNN~\cite{wang2021crgnn}, includes subsets \textbf{V-HIM2K5} for training and \textbf{V-HIM60} for testing. We categorized the dataset into three difficulty levels based on instance overlap. \Tref{tab:data_vid_info} shows some details of the synthesized datasets. Masks in training involved dilation and erosion on binarized alpha mattes. For testing, masks are generated using XMem~\cite{cheng2022xmem}. Further details on dataset synthesis and difficulty levels are provided in the supplementary material.

\section{Experiments}

We developed our model using PyTorch~\cite{imambi2021pytorch} and the Sparse convolution library Spconv~\cite{spconv2022}. Our codebase is built upon the publicly available implementations of MGM~\cite{yu2021mgm} and OTVM~\cite{seong2022otvm}. In the first~\Sref{sec:pretrain}, we discuss the results when pre-training on the image matting dataset. The performance on the video dataset is shown in the~\Sref{sec:train_video}. All training settings are reported in the supplementary material.

\subsection{Pre-training on image data}
\label{sec:pretrain}

\begin{table}[t!]
    \vspace{-1.5em}
    \centering
    \caption{\textbf{Superiority of Mask Embedding Over Stacking in HIM2K+M-HIM2K.} Our mask embedding technique demonstrates enhanced performance compared to traditional stacking methods.}
    \small
    \begin{tblr}{width=\linewidth,colsep=3.8pt,colspec={@{}l|ccc|ccc@{}}}
    \toprule
         \SetCell[r=2]{c}{Mask input}
                & \SetCell[c=3]{c}{Composition} & & 
                    & \SetCell[c=3]{c}{Natural} \\
    \hline
    & MAD & Grad & Conn & MAD & Grad & Conn \\
    \hline
    Stacked & 27.01 & 16.80 & 15.72 
        & 39.29 & 16.44 & 23.26 \\ 
    Embeded($C_e=1$) & 19.18 & 13.00 & 11.16 
        & 33.60 & 13.44 & 19.18 \\ 
    Embeded($C_e=2$) & 21.74 & 14.39 & 12.69 
        & 35.16 & 14.51 & 20.40 \\ 
    Embeded($C_e=3$) & \textbf{17.75} & \textbf{12.52} & \textbf{10.32} 
        & \textbf{33.06} & \textbf{13.11} & \textbf{17.30} \\ 
    Embeded($C_e=5$) & 24.79 & 16.19 & 14.58 
        & 34.25 & 15.66 & 19.70 \\
    \bottomrule
    \end{tblr}
    \label{tab:abl_mask_inp}
    \vspace{-1.5em}
\end{table}

\begin{table}[b]
    \vspace{-1em}
    \centering
    \small
    \caption{\textbf{Optimal Performance with $\mathcal{L}_{att}$ and $\mathbf{W}8$ on HIM2K+M-HIM2K.} Utilizing both $\mathcal{L}_{att}$ and $\mathbf{W}_8$ leads to superior results.}
    \vspace{-0.5em}
    \begin{tblr}{width=\linewidth,colsep=5.5pt,colspec={@{}cc|ccc|ccc@{}}}
         \toprule
         \SetCell[r=2]{c}{$\mathcal{L}_{att}$} 
                & \SetCell[r=2]{c}{$\mathbf{W}_8$}
                & \SetCell[c=3]{c}{Composition} & & 
                    & \SetCell[c=3]{c}{Natural} \\
    \hline
    & &  MAD & Grad & Conn & MAD & Grad & Conn \\
    \hline
    &  & 
        31.77 & 16.58 & 18.27 & 46.68 & 15.68 & 30.64  \\
    & \checkmark & 
        25.41 & 14.53 & 14.75 & 46.30 & 15.84 & 29.26 \\
    \checkmark &  & 
        17.56 & \textbf{12.34} & 10.22 & 32.95 & 13.29 & \textbf{17.06} \\
    \checkmark & \checkmark & 
        \textbf{17.55} & \textbf{12.34} & \textbf{10.19} & \textbf{32.03} & \textbf{13.16} & 17.43 \\
    \bottomrule
    \end{tblr}
    
    \label{tab:abl_loss}
\end{table}

\begin{table*}[t!]
    \vspace{-1.5em}
    \centering
    \caption{\textbf{Comparative Performance on HIM2K+M-HIM2K.} Our method outperforms baselines, with average results (large numbers) and standard deviations (small numbers) on the benchmark. The upper group represents methods predicting each instance separately, while the lower models utilize instance information. Gray rows denote public weights trained on external data, not retrained on I-HIM50K. MGM$^\dagger$ denotes the MGM-in-the-wild. MGM$^\star$ refers to MGM with all masks stacked with the input image. Models are tested on images with a short side of 576px. \fst{Bold} and \snd{underline} highlight the best and second-best models per metric, respectively.}
    \vspace{-0.5em}
    \footnotesize %
    \begin{tblr}{width=\textwidth,colsep=1mm,colspec={@{}l|cccccc|cccccc@{}},row{4,8}={gainsboro}}
    \toprule
    \SetCell[r=2]{l} Method
         & \SetCell[c=6]{c} Composition set & & & & &
         & \SetCell[c=6]{c} Natural set \\
    \hline
    & MAD & MSE & Grad & Conn & MAD$_f$  & MAD$_u$ 
    & MAD & MSE & Grad & Conn & MAD$_f$  & MAD$_u$ \\
    \hline
    \SetCell[c=13]{l} \textbf{\textit{Instance-agnostic}} \\
    \hline
    MGM$^\dagger$~\cite{park2023mgmwild} & 23.15 \std{1.5} & 14.76 \std{1.3} & 12.75 \std{0.5} & 13.30 \std{0.9} & 64.39 \std{4.5} & 309.38 \std{12.0} 
        & 32.52 \std{6.7}  & 18.80 \std{6.0} & 12.52 \std{1.2} & 18.51 \std{18.5} & 65.20 \std{15.9} & 179.76 \std{23.9} \\ 
    MGM~\cite{yu2021mgm} & 15.32 \std{0.6} & 9.13 \std{0.5} & \snd{9.94 \std{0.2}} & 8.83 \std{0.3} & \snd{33.54 \std{1.9}}  & 261.43 \std{4.0}
        & 30.23 \std{3.6} & 17.40 \std{3.3} & \snd{10.53 \std{0.5}} & 15.70 \std{1.9} & 63.16 \std{13.0} & 167.35 \std{12.1} \\ 
    SparseMat~\cite{sun2023sparsemat} & 21.05 \std{1.2} & 14.55 \std{1.0} & 14.64 \std{0.5} & 12.26 \std{0.7} & 45.19 \std{2.9} & 352.95 \std{14.2} 
        & 35.03 \std{5.1} & 21.79 \std{4.7} & 15.85 \std{1.2} & 18.50 \std{3.1} & 67.82 \std{15.2} & 212.63 \std{20.8} \\ 
    \hline
    \SetCell[c=13]{l} \textbf{\textit{Instance-aware}} \\
    \hline
    InstMatt~\cite{sun2022instmatt} & 12.85 \std{0.2} & 5.71 \std{0.2} & 9.41 \std{0.1} & 7.19 \std{0.1} & 22.24 \std{1.3} & 255.61 \std{2.0} 
        & 26.76 \std{2.5} & 12.52 \std{2.0} & 10.20 \std{0.3} & 13.81 \std{1.1} & 48.63 \std{6.8} & 161.52 \std{6.9} \\ 
    InstMatt~\cite{sun2022instmatt} & 16.99 \std{0.7} & 9.70 \std{0.5} & 10.93 \std{0.3} & 9.74 \std{0.5} & 53.76 \std{3.0} & 286.90 \std{7.0} 
        & \snd{28.16 \std{4.5}} & \fst{14.30 \std{3.7}} & 10.98 \std{0.7} & \snd{14.63 \std{2.0}} & 57.83 \std{12.1} & 168.74 \std{15.5} \\ 
    MGM$^\star$ & \snd{14.31 \std{0.4}} & \snd{7.89 \std{0.4}} & 10.12 \std{0.2} & \snd{8.01 \std{0.2}} & 41.94 \std{3.1} & \snd{251.08 \std{3.6}}
        & 31.38 \std{3.3} & 18.38 \std{3.1} & 10.97 \std{0.4} & 14.75 \std{1.4} & \snd{53.89 \std{9.6}} & \snd{165.13 \std{10.6}} \\
    MaGGIe (ours) & \fst{12.93 \std{0.3}} & \fst{7.26 \std{0.3}} & \fst{8.91 \std{0.1}} & \fst{7.37 \std{0.2}} & \fst{19.54 \std{1.0}} & \fst{235.95 \std{3.4}}
        & \fst{27.17 \std{3.3}} & \snd{16.09 \std{3.2}} & \fst{9.94 \std{0.6}} & \fst{13.42 \std{1.4}} & \fst{49.52 \std{8.0}} & \fst{146.71 \std{11.6}} \\
    \bottomrule
    \end{tblr}
    \label{tab:image_matting}
    \vspace{-1.0em}
\end{table*}

\mypara{Metrics.} Our evaluation metrics included Mean Absolute Differences (MAD), Mean Squared Error (MSE), Gradient (Grad), and Connectivity (Conn). We also separately computed these metrics for the foreground and unknown regions, denoted as MAD$_f$ and MAD$_u$, by estimating the trimap on the ground truth. Since our images contain multiple instances, metrics were calculated for each instance individually and then averaged. We did not use the IMQ from InstMatt, as our focus is not on instance detection.

\mypara{Ablation studies.} Each ablation study setting was trained for 10,000 iterations with a batch size 96. We first assessed the performance of the embedding layer versus stacked masks and image inputs in~\Tref{tab:abl_mask_inp}. The mean results on M-HIM2K are reported, with full results in the supplementary material. The embedding layer showed improved performance, particularly effective with $C_e=3$. We also evaluated the impact of using $\mathcal{L}_{att}$ and $\mathbf{W}_8$ in training in~\Tref{tab:abl_loss}. $\mathcal{L}_{att}$ significantly enhanced model performance, while $\mathbf{W}_8$ provided a slight boost.

\begin{figure}[b]
    \centering
    \vspace{-1em}
    \includegraphics[width=0.49\columnwidth]{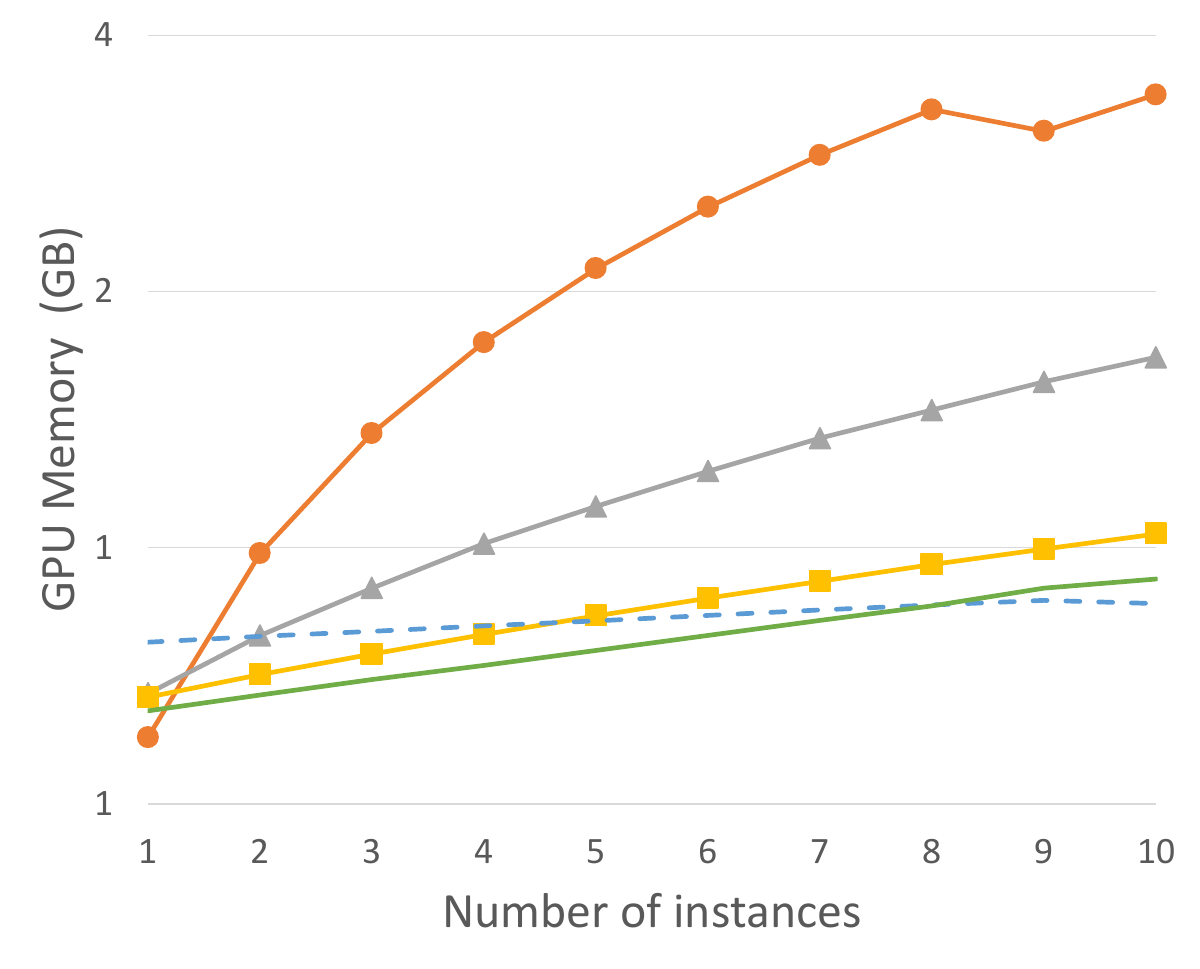}
    \includegraphics[width=0.49\columnwidth]{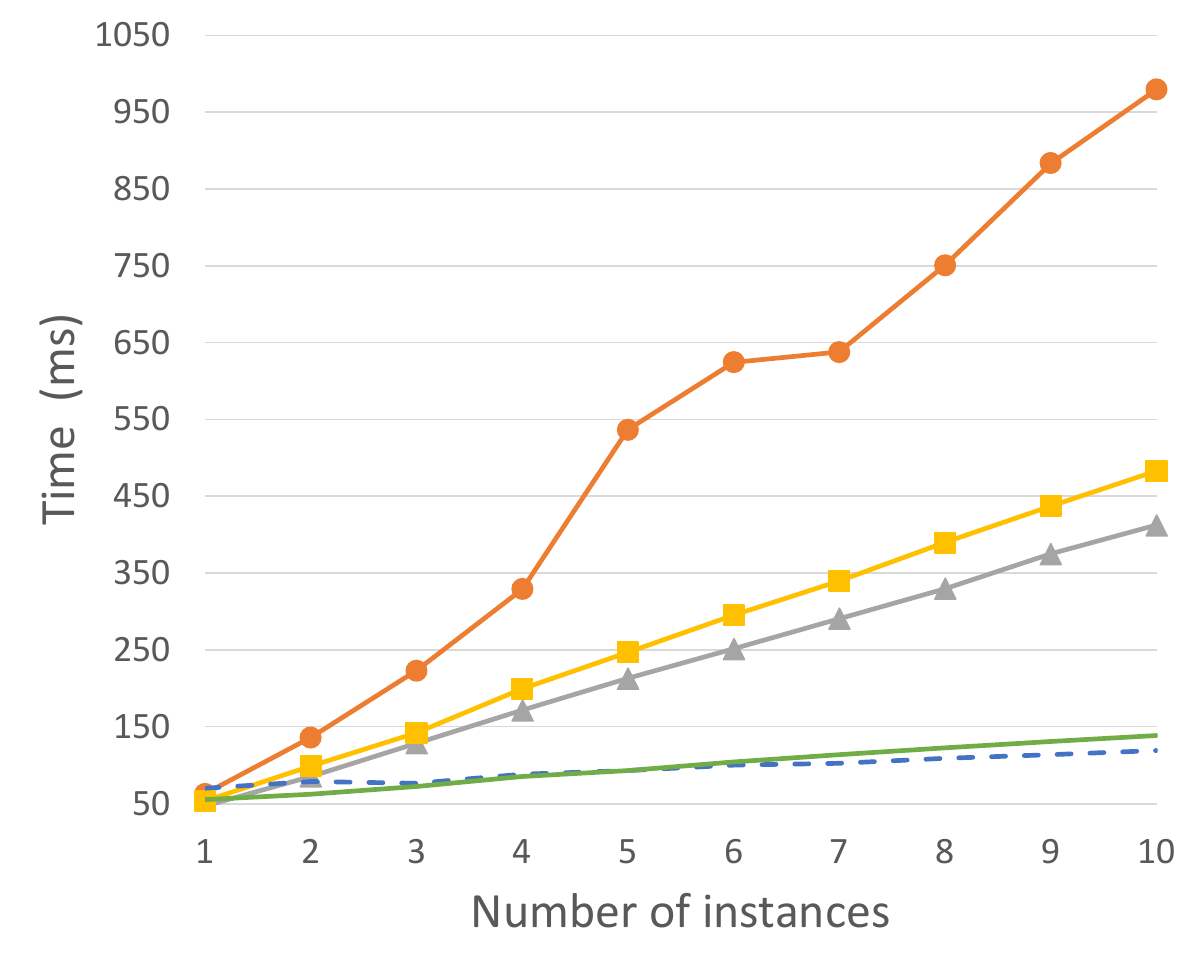}
    \includegraphics[width=0.75\columnwidth]{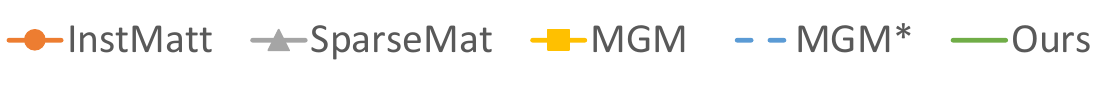}
    \vspace{-0.5em}
    \caption{\textbf{Our model keeps steady memory and time complexity when the number of instance increases.} InstMatt's complexity increases linearly with the number of instances.}
    \label{fig:mem_time}
    \vspace{-1.5em}
\end{figure}

\mypara{Quantitative results.} We evaluated our model against previous baselines after retraining them on our I-HIM50K dataset. Besides original works, we modified SparseMat's first layer to accept a single mask input. Additionally, we expanded MGM to handle up to 10 instances, denoted as MGM$^\star$. We also include the public weights of InstMatt~\cite{sun2022instmatt} and MGM-in-the-wild~\cite{park2023mgmwild}. 
The performance with different masks M-HIM2K are reported in~\Tref{tab:image_matting}. The public InstMatt showed the best performance, but this comparison may not be entirely fair as it was trained on private external data. Our model demonstrated comparable results on composite and natural sets, achieving the lowest error in most metrics. MGM$^\star$ also performed well, suggesting that processing multiple masks simultaneously can facilitate instance interaction, although this approach slightly impacted the Grad metric, which reflects the output's detail. 

We also measure the memory and speed of models on M-HIM2K natural set in~\Fref{fig:mem_time}. While InstMatt, MGM, and SparseMat have the inference time increasing linearly to the number of instances, MGM$^\star$ and ours keep steady performance in both memory and speed.

\begin{table}[b]
    \vspace{-1em}
    \centering
    \caption{\textbf{Superiority of Temporal Consistency in Feature and Prediction Levels.} Our MaGGIe, integrating temporal consistency at both feature and matte levels, outperforms non-temporal methods and those with only feature level.}
    \vspace{-0.5em}
    \footnotesize
    \begin{tblr}{width=\columnwidth,colsep=3.5pt,colspec={@{}cc|cc|cc|cc|cc@{}}}
    \toprule
         \SetCell[c=2]{c}{Conv-GRU} & 
        & \SetCell[c=2]{c}{Fusion} & 
        & \SetCell[c=2]{c}{Easy} &
        & \SetCell[c=2]{c}{Medium} &
        & \SetCell[c=2]{c}{Hard} &\\
    \hline
    Single & Bi & $\mathbf{\hat{A}}^f$ &  $\mathbf{\hat{A}}^b$ & MAD & dtSSD & MAD & dtSSD & MAD & dtSSD \\
    \hline
    & & &                                   & 10.26 & 16.57 & 13.88 & 23.67 & 21.62 & 30.50 \\ 
    \hline
    \checkmark & & &                        & 10.15 & 16.42 & 13.84 & 23.66 & 21.26 & 29.95 \\ 
    & \checkmark & &                        & 10.14 & 16.41 & \textbf{13.83} & 23.66 & 21.25 & 29.92 \\ 
    & \checkmark & \checkmark &             & 11.32 & 16.51 & 15.33 & 24.08 & 24.97 & 30.66 \\ 
    & \checkmark & \checkmark & \checkmark  & \textbf{10.12} & \textbf{16.40} & 13.85 & \textbf{23.63} & \textbf{21.23} & \textbf{29.90} \\
    \bottomrule
    \end{tblr}
    \label{tab:abl_temp}
    \vspace{-1.5em}
\end{table}

\begin{figure*}[ht!]
    \centering
    \vspace{-1.5em}
    \includegraphics[width=\textwidth]{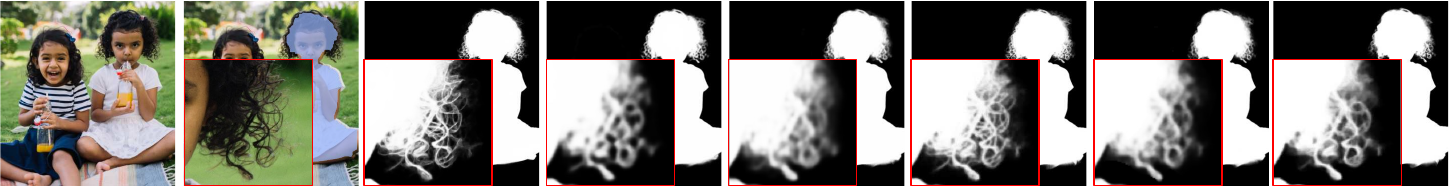}
    \includegraphics[width=\textwidth]{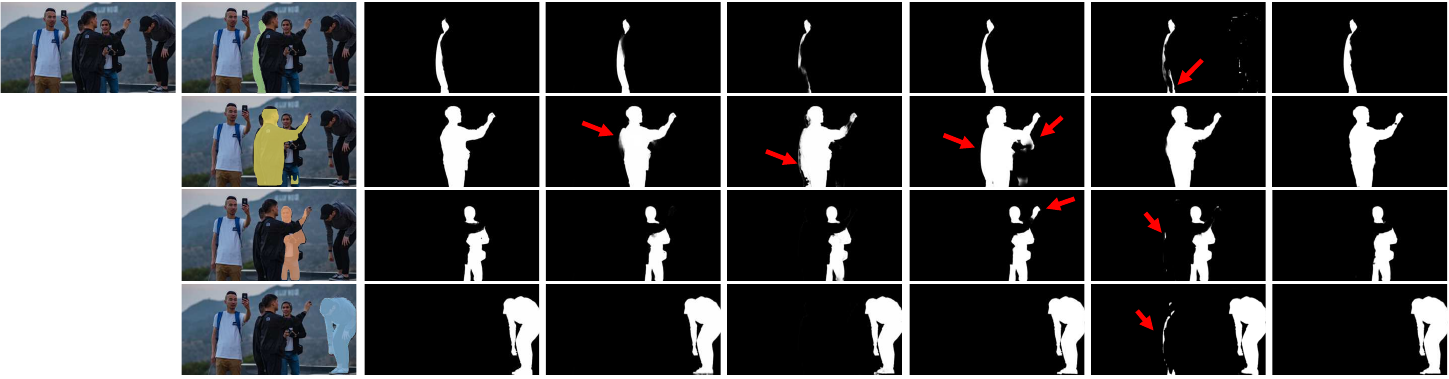}
    \begin{tblr}{width=\textwidth,colspec={X[1,c]X[1,c]X[1,c]X[1,c]X[1,c]X[1,c]X[1,c]X[1,c]}}
        Image & Input mask & Groundtruth & \colorbox{gainsboro}{InstMatt} & InstMatt & MGM & MGM$^\star$ & Ours
    \end{tblr}
    \vspace{-1em}
    \caption{\textbf{Enhanced Detail and Instance Separation by MaGGIe.} Our model excels in rendering detailed outputs and effectively separating instances, as highlighted by {\color{red} red} squares (detail focus) and {\color{red}red} arrows (errors in other methods).}
    \label{fig:qual_image}
\end{figure*}

\begin{table*}[ht!]
    \centering
    \caption{\textbf{Comparative Analysis of Video Matting Methods on V-HIM60}. This table categorizes methods into two groups: those utilizing first-frame trimaps (upper group) and mask-guided approaches (lower group). Gray rows denotes models with public weights not retrained on I-HIM50K and V-HIM50K. MGM$^\star$-TCVOM represents MGM with stacked guidance masks and the TCVOM temporal module. \fst{Bold} and \snd{underline} highlight the top and second-best performing models in each metric, respectively.}
    \vspace{-0.5em}
    \footnotesize
     \begin{tblr}{width=\textwidth,colsep=3pt,colspec={@{}l|rrrrr|rrrrr|rrrrr@{}},row{4,6}={gainsboro}}
     \toprule
        \SetCell[r=2]{l} Method 
            & \SetCell[c=5]{c} Easy 
            & & & & & \SetCell[c=5]{c} Medium
            & & & & & \SetCell[c=5]{c} Hard \\
    \hline
    & MAD & Grad & Conn & dtSSD & \scriptsize{MESSDdt} & MAD & Grad & Conn & dtSSD & \scriptsize{MESSDdt} & MAD & Grad & Conn & dtSSD & \scriptsize{MESSDdt} \\
    \hline
    \SetCell[c=16]{l}{\textbf{\textit{First-frame trimap}}} \\
    \hline
    OTVM~\cite{seong2022otvm} & 204.59 & 15.25 & 76.36 & 46.58 & 397.59 
        & 247.97 & 21.02 & 97.74 & 66.09 & 587.47 
        & 412.41 & 29.97 & 146.11 & 90.15 & 764.36 \\ 
    OTVM~\cite{seong2022otvm} & 36.56 & 6.62 & 14.01 & 24.86 & 69.26 
        & 48.59 & 10.19 & 17.03 & 36.06 & 80.38 
        & 140.96 & 17.60 & 47.84 & 59.66 & 298.46 \\ 
    FTP-VM~\cite{huang2023ftpvm} & 12.69 & 6.03 & 4.27 & 19.83 & 18.77 
        & 40.46 & 12.18 & 15.13 & 32.96 & 125.73 
        & 46.77 & 14.40 & 15.82 & 45.04 & 76.48 \\ 
    FTP-VM~\cite{huang2023ftpvm} & 13.69 & 6.69 & 4.78 & 20.51 & 22.54 
        & 26.86 & 12.39 & 9.95 & 32.64 & 126.14 
        & 48.11 & 14.87 & 16.12 & 45.29 & 78.66 \\ 
    \hline
    \SetCell[c=16]{l}{\textbf{\textit{Frame-by-frame binary mask}}} \\
    \hline
    MGM-TCVOM~\cite{seong2022otvm} & 11.36 & 4.57 & 3.83 & 17.02 & 19.69 
        & 14.76 & 7.17 & 5.41 & \fst{23.39} & \snd{39.22}
        & \snd{22.16} & 7.91 & \snd{7.27} & \snd{31.00} & 47.82 \\
    MGM$^\star$-TCVOM~\cite{seong2022otvm} & \snd{10.97} & \snd{4.19} & \snd{3.70} & \snd{16.86} & \fst{15.63}
        & \fst{13.76} & \snd{6.47} & \fst{5.02} & 23.99 & 42.71 
        & 22.59 & \snd{7.86} & 7.32 & 32.75 & \fst{37.83} \\
    InstMatt~\cite{sun2022instmatt} & 13.77 & 4.95 & 3.98 & 17.86 & 18.22 
        & 19.34 & 7.21 & 6.02 & 24.98 & 54.27 
        & 27.24 & 7.88 & 8.02 & 31.89 & 47.19 \\
    SparseMat~\cite{sun2023sparsemat} & 12.02 & 4.49 & 4.11 & 19.86 & 24.75 
        & 18.20 & 8.03 & 6.87 & 30.19 & 85.79 
        & 24.83 & 8.47 & 8.19 & 36.92 & 55.98 \\
    MaGGIe (ours) & \fst{10.12} & \fst{4.08} & \fst{3.43} & \fst{16.40} & \snd{16.41}
        & \snd{13.85} & \fst{6.31} & \snd{5.11} & \snd{23.63} & \fst{38.12}
        & \fst{21.23} & \fst{7.08} & \fst{6.89} & \fst{29.90} & \snd{42.98} \\
    \bottomrule
    \end{tblr}
    
    \label{tab:video_matting}
    \vspace{-1.0em}
\end{table*}

\mypara{Qualitative results.} MaGGIe's ability to capture fine details and effectively separate instances is showcased in~\Fref{fig:qual_image}. At the exact resolution, our model not only achieves highly detailed outcomes comparable to running MGM separately for each instance but also surpasses both the public and retrained versions of InstMatt.
A key strength of our approach is its proficiency in distinguishing between different instances. This is particularly evident when compared to MGM, where we observed overlapping instances, and MGM$^\star$, which has noise issues caused by processing multiple masks simultaneously. Our model's refined instance separation capabilities highlight its effectiveness in handling complex matting scenarios.

\subsection{Training on video data}
\label{sec:train_video}

\mypara{Temporal consistency metrics.} Following previous works~\cite{zhang2021tcvom,sun2021dvm,seong2022otvm}, we extended our evaluation metrics to include dtSSD and MESSDdt to assess the temporal consistency of instance matting across frames.

\mypara{Ablation studies. } Our tests, detailed in \Tref{tab:abl_temp}, show that each temporal module significantly impacts performance. Omitting these modules increased errors in all subsets. Single-direction Conv-GRU use improved outcomes, with further gains from adding backward pass fusion. Forward fusion alone was less effective, possibly due to error propagation. The optimal setup involved combining backward propagation to reduce errors, yielding the best results. 

\begin{figure*}[ht!]
    \centering
    \vspace{-1em}
    \footnotesize
    \includegraphics[width=\textwidth]{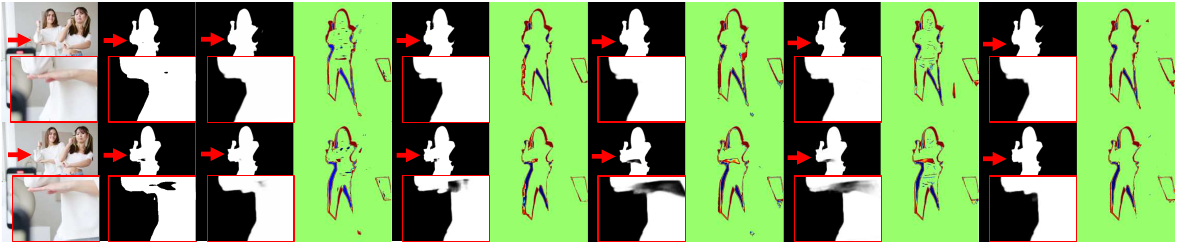}
    \includegraphics[width=\textwidth]{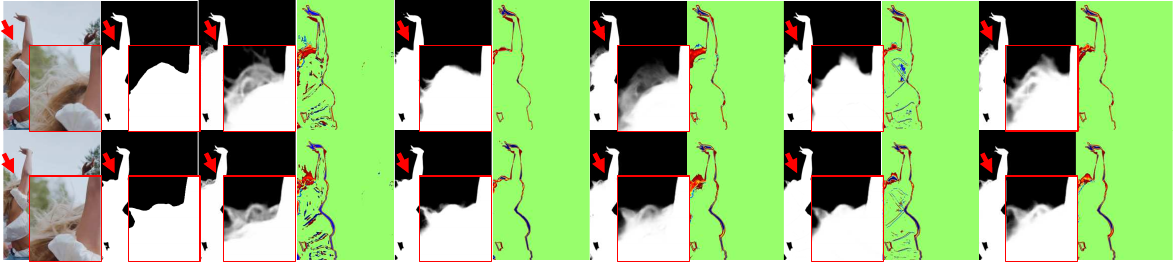}
    \begin{tblr}{width=\textwidth,rowsep=-6pt,colsep=3pt,colspec={X[1,c]X[1,c]X[1,c]X[1,c]X[1,c]X[1,c]X[1,c]X[1,c]X[1,c]X[1,c]X[1,c]X[1,c]X[1,c]}}
        \SetCell[r=2]{c}{Frame} & \SetCell[r=2]{c}{Input mask} & 
            $\hat{\mathbf{A}}$ & $\log\lft|\Delta_{\hat{\mathbf{A}}}\rgt|$ & 
            $\hat{\mathbf{A}}$ & $\log\lft|\Delta_{\hat{\mathbf{A}}}\rgt|$ & 
            $\hat{\mathbf{A}}$ & $\log\lft|\Delta_{\hat{\mathbf{A}}}\rgt|$ & 
            $\hat{\mathbf{A}}$ & $\log\lft|\Delta_{\hat{\mathbf{A}}}\rgt|$ & 
            $\hat{\mathbf{A}}$ & $\log\lft|\Delta_{\hat{\mathbf{A}}}\rgt|$ \\
        \vspace{-5cm}
        & & \SetCell[c=2]{c}{$\underbrace{\hspace{2.6cm}}_{\substack{\vspace{-5.0mm}}\colorbox{white}{~~InstMatt~~}}$}  &
        & \SetCell[c=2]{c}{$\underbrace{\hspace{2.6cm}}_{\substack{\vspace{-5.0mm}}\colorbox{white}{~~SparseMat~~}}$} &
        & \SetCell[c=2]{c}{$\underbrace{\hspace{2.6cm}}_{\substack{\vspace{-5.0mm}}\colorbox{white}{~~MGM-TCVOM~~}}$} &
        & \SetCell[c=2]{c}{$\underbrace{\hspace{2.6cm}}_{\substack{\vspace{-5.0mm}}\colorbox{white}{~~MGM$^\star$-TCVOM~~}}$} &
        & \SetCell[c=2]{c}{$\underbrace{\hspace{2.6cm}}_{\substack{\vspace{-5.0mm}}\colorbox{white}{~~MaGGIe (ours)~~}}$} \\
    \end{tblr}
    \caption{\textbf{Detail and Consistency in Frame-to-Frame Predictions}. This figure demonstrates the precision and temporal consistency of our model's alpha matte predictions, highlighting robustness against noise from input masks. The color-coded map (min-max range) to illustrate differences between consecutive frames is \includegraphics[height=8pt]{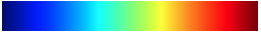}.}
    \label{fig:qual_vid}
    \vspace{-1.5em}
\end{figure*}

\mypara{Performance evaluation.} Our model was benchmarked against leading methods in trimap video matting, mask-guided matting, and instance matting. For trimap video matting, we chose OTVM~\cite{seong2022otvm} and FTP-VM~\cite{huang2023ftpvm}, fine-tuning them on our V-HIM2K5 dataset. In masked guided video matting, we compared our model with InstMatt~\cite{sun2022instmatt}, SparseMat~\cite{sun2023sparsemat}, and MGM~\cite{yu2021mgm} which is combined with the TCVOM~\cite{zhang2021tcvom} module for temporal consistency. InstMatt, after being fine-tuned on I-HIM50K and subsequently on V-HIM2K5, processed each frame in the test set independently, without temporal awareness. SparseMat, featuring a temporal sparsity fusion module, was fine-tuned under the same conditions as our model. MGM and its variant, integrated with the TCVOM module, emerged as strong competitors in our experiments, demonstrating their robustness in maintaining temporal consistency across frames.

The comprehensive results of our model across three test sets, using masks from XMem, are detailed in~\Tref{tab:video_matting}. All trimap propagation methods are underperform the mask-guided solutions. When benchmarked against other masked guided matting methods, our approach consistently reduces error across most metrics. Notably, it excels in temporal consistency, evidenced by its top performance in dtSSD for both easy and hard test sets, and in MESSDdt for the medium set. Additionally, our model shows superior performance in capturing fine details, as indicated by its leading scores in the Grad metric across all test sets. These results underscore our model's effectiveness in video instance matting, particularly in challenging scenarios requiring high temporal consistency and detail preservation.

\mypara{Temporal consistency and detail preservation.} Our model's effectiveness in video instance matting is evident in~\Fref{fig:qual_vid} with natural videos. Key highlights include:

\begin{itemize}
    \item \textit{Handling of Random Noises}: Our method effectively handles random noise in mask inputs, outperforming others that struggle with inconsistent input mask quality.
    \item \textit{Foreground/Background Region Consistency}: We maintain consistent, accurate foreground predictions across frames, surpassing InstMatt and MGM$^\star$-TCVOM.
    \item \textit{Detail Preservation}: Our model retains intricate details, matching InstMatt's quality and outperforming MGM variants in video inputs.
\end{itemize}

These aspects underscore MaGGIe's robustness and effectiveness in video instance matting, particularly in maintaining temporal consistency and preserving fine details across frames.

\section{Discussion}

\mypara{Limitation and Future work.} Our MaGGIe demonstrates effective performance in human video instance matting with binary mask guidance, yet it also presents opportunities for further research and development. One notable limitation is the reliance on one-hot vector representation for each location in the guidance mask, necessitating that each pixel is distinctly associated with a single instance. This requirement can pose challenges, particularly when integrating instance masks from varied sources, potentially leading to misalignments in certain regions. Additionally, the use of composite training datasets may constrain the model's ability to generalize effectively to natural, real-world scenarios. While the creation of a comprehensive natural dataset remains a valuable goal, we propose an interim solution: the utilization of segmentation datasets combined with self-supervised or weakly-supervised learning techniques. This approach could enhance the model's adaptability and performance in more diverse and realistic settings, paving the way for future advancements in the field.

\mypara{Conclusion.} Our study contributes to the evolving field of instance matting, with a focus that extends beyond human subjects. By integrating advanced techniques like transformer attention and sparse convolution, MaGGIe shows promising improvements over previous methods in detailed accuracy, temporal consistency, and computational efficiency for both image and video inputs. Additionally, our approach in synthesizing training data and developing a comprehensive benchmarking schema offers a new way to evaluate the robustness and effectiveness of models in instance matting tasks. This work represents a step forward in video instance matting and provides a foundation for future research in this area.

\mypara{Acknownledgement.} We sincerely appreciate Markus Woodson for the invaluable initial discussions. Additionally, I am deeply thankful to my wife, Quynh Phung, for her meticulous proofreading and feedback.

{
    \small
    \bibliographystyle{ieeenat_fullname}
    \bibliography{main}
}

\clearpage
\maketitlesupplementary

\addtocontents{toc}{\protect\setcounter{tocdepth}{2}}
\let\oldaddcontentsline\addcontentsline
\def\addcontentsline#1#2#3{}

\def\addcontentsline#1#2#3{\oldaddcontentsline{#1}{#2}{#3}}
\begingroup
\let\clearpage\relax
{
  \hypersetup{linkcolor=cvprblue}
  \tableofcontents
}
\endgroup
\section{Architecture details}
This section delves into the architectural nuances of our framework, providing a more detailed exposition of components briefly mentioned in the main paper. These insights are crucial for a comprehensive understanding of the underlying mechanisms of our approach.

\subsection{Mask guidance identity embedding} 
\label{sec:sup_idembedding}
We embed mask guidance into a learnable space before inputting it into our network. This approach, inspired by the ID assignment in AOT~\cite{yang2021aot}, generates a guidance embedding $\mathbf{E} \in \R^{T\x C_e \x H \x W}$ by mapping embedding vectors $\mathbf{D}\in \R^{N \x C_e}$ to pixels based on the guidance mask $\mathbf{M}$:
\begin{equation}
    \mathbf{E}(x,y) = \mathbf{M}(x,y) \mathbf{D}.
\end{equation}
Here, $\mathbf{E}(x,y) \in \R^{T\x C_e}$ and $\mathbf{M}(x,y) \in \{0,1\}^{T\x N}$ represent the values at row $y$ and column $x$ in $\mathbf{E}$ and $\mathbf{M}$, respectively. In our experiment, we set $N=10$, but it can be any larger number without affecting the architecture significantly.

\subsection{Feature extractor}
\label{sec:sup_backbone}
In our experiments, we employ ResNet-29~\cite{he2016deep} as the feature extractor, consistent with other baselines~\cite{yu2021mgm,sun2022instmatt}. We have $C_8=128,C_4=64,C_1=C_2=32$.

\subsection{Dense-image to sparse-instance features}
\label{sec:sup_dense2sparse}

\begin{figure}[t]
    \centering
    \includegraphics[width=\columnwidth]{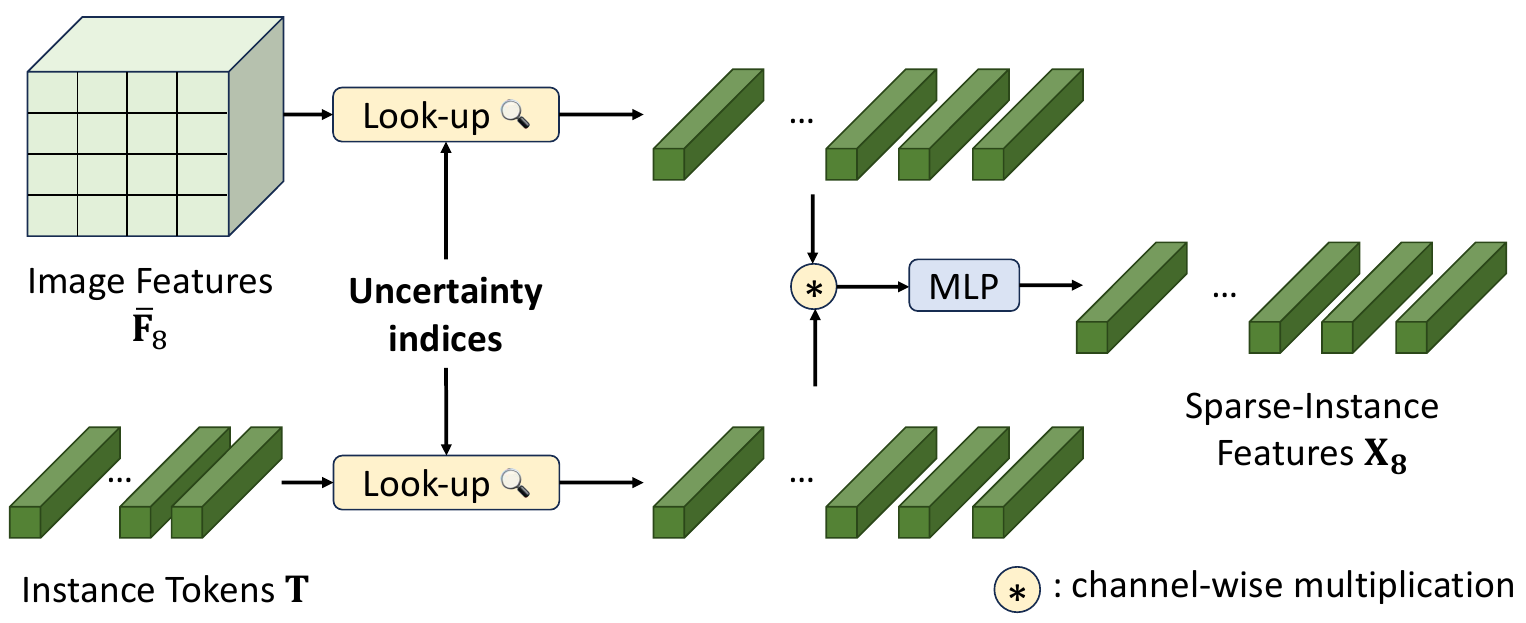}
    \caption{\textbf{Converting Dense-Image to Sparse-Instance Features}. We transform the dense image features into sparse, instance-specific features with the help of instance tokens.}
    \label{fig:method_di2si}
\end{figure}

We express the~\Eref{eq:dense2sparse} as the visualization in~\Fref{fig:method_di2si}. It involves extracting feature vectors $\mathbf{\bar{F}}(x,y,t)$ and instance token vectors $\mathbf{T}_i$ for each uncertainty index $(x,y,t,i)\in \mathbf{U}$. These vectors undergo channel-wise multiplication, emphasizing channels relevant to each instance. A subsequent MLP layer then converts this product into sparse, instance-specific features.

\subsection{Detail aggregation}
\label{sec:sup_detail_aggre}

\begin{figure}[b]
    \centering
    \includegraphics[width=\columnwidth]{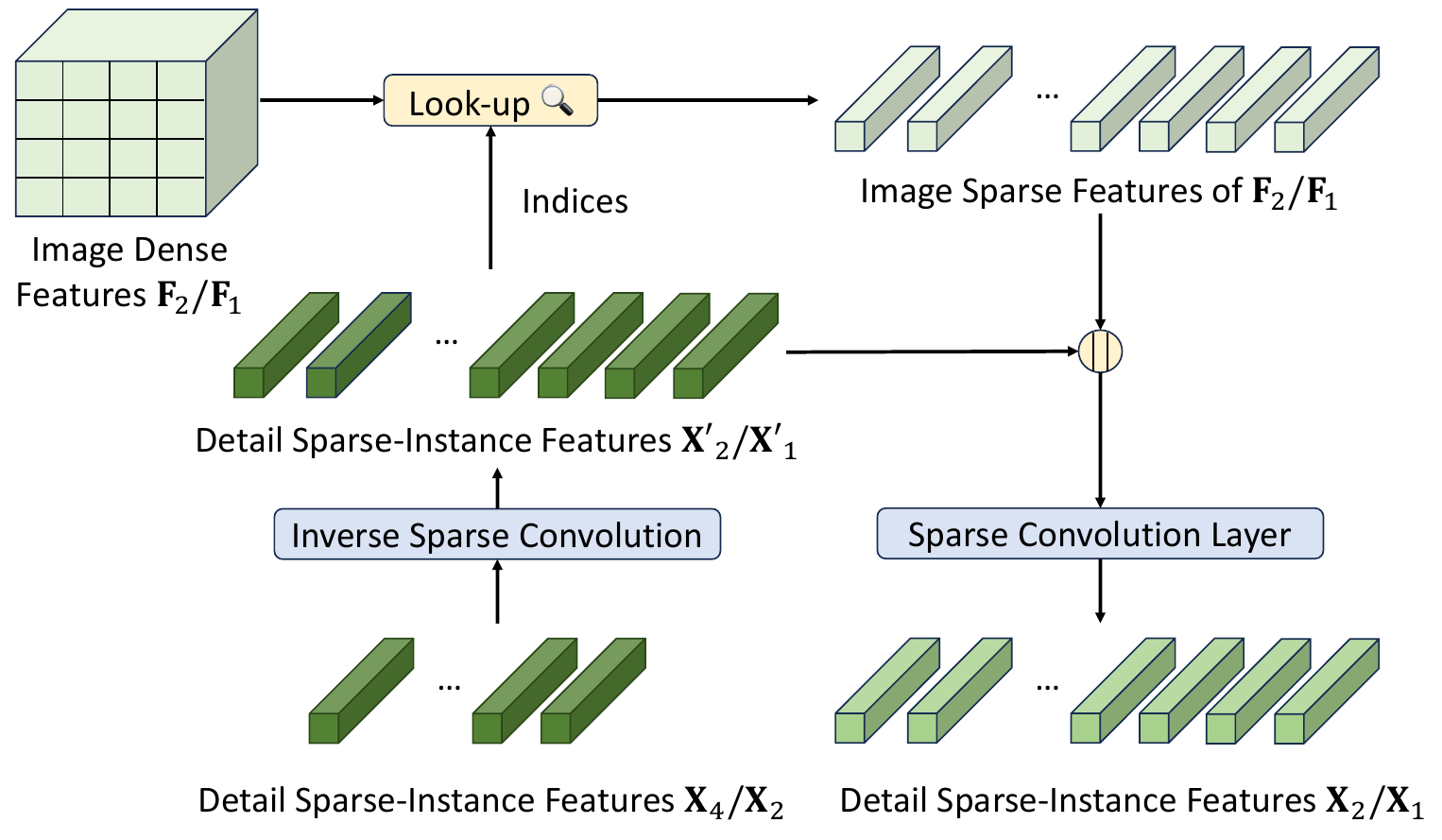}
    \caption{\textbf{Detail Aggregation Module merges sparse features across scales}. This module equalizes spatial scales of sparse features using inverse sparse convolution, facilitating their combination.}
    \label{fig:method_sdam}
\end{figure}

This process, akin to a U-net decoder, aggregates features from different scales, as detailed in~\Fref{fig:method_sdam}. It involves upscaling sparse features and merging them with corresponding higher-scale features. However, this requires pre-computed downscale indices from dummy sparse convolutions on the full input image.

\subsection{Sparse matte head}
\label{sec:sup_sparse_matte_head}
Our matte head design, inspired by MGM~\cite{yu2021mgm}, comprises two sparse convolutions with intermediate normalization and activation (Leaky ReLU) layers. The final output undergoes sigmoid activation for the final prediction. Non-refined locations in the dense prediction are assigned a value of zero.

\subsection{Sparse progressive refinement}
\label{sec:sup_prm}
The PRM module progressively refines $\mathbf{A}_8 \rightarrow \mathbf{A}_4 \rightarrow \mathbf{A}_1$ to have $\mathbf{A}$. We assume that all predictions are rescaled to the largest size and perform refinement between intermediate predictions and uncertainty indices $\mathbf{U}$:
\begin{align}
    \mathbf{A} &= \mathbf{A}_8 \\
    \mathbf{R}_4(j) &= \begin{cases}
        1 \text{, if } j \in \mathcal{D}(\mathbf{A}) \text{ and } j \in \mathbf{U} \\
        0 \text{, otherwise }
    \end{cases}    \\
    \mathbf{A} &= \mathbf{A} \x (1 - \mathbf{R}_4) + \mathbf{R}_4 \x \mathbf{A}_4 \\
    \mathbf{R}_1(j) &= \begin{cases}
        1 \text{, if } j \in \mathcal{D}(\mathbf{A}) \text{ and } j \in \mathbf{U} \\
        0 \text{, otherwise }
    \end{cases}    \\
    \mathbf{A} &= \mathbf{A} \x (1 - \mathbf{R}_1) + \mathbf{R}_1 \x \mathbf{A}_4
\end{align}
where $j=(x,y,t,i)$ is an index in the output; $\mathbf{R}_1, \mathbf{R}_4$ in shape $T \x N \x H \x W$; and $\mathcal{D}(\mathbf{A}) = \text{dilation}(0 < \mathbf{A} < 1)$ is the indices of all dilated uncertainty values on $\mathbf{A}$. The dilation kernel is set to 30, 15 for $\mathbf{R}_4, \mathbf{R}_1$ respectively.

\subsection{Attention loss and loss weight}
\label{sec:sup_loss}
With $\mathbf{A}^{gt}$ as the ground-truth alpha matte and its $\frac{1}{8}$ downscaled version $\mathbf{A}^{gt}_8$, we define a binarized $\mathbf{\tilde{A}}^{gt}_8 = \mathbf{A}^{gt}_8 > 0$. The attention loss $\mathcal{L}_{\text{att}}$ is:
\begin{equation}
    \mathcal{L}_{\text{att}} = \sum_i^{N} \left\|\mathbf{1} -  \text{Aff}(i)^\top \mathbf{\tilde{A}}^{gt}_8(i) \right\|_1
\end{equation}
aiming to maximize each instance token $\mathbf{T}_i$'s attention score to its corresponding groundtruth region $\mathbf{\tilde{A}}^{gt}_8(i)$.

The weight $\mathbf{W}_8$ at each location is:
\begin{equation}
    \mathbf{W}_8(j) = \begin{cases}
        \gamma , \text{ if } 0 < \mathbf{A}^{gt}_8(j) < 1 \text{ and } 0 < \mathbf{A}_8(j) < 1 \\
        1.0, \text{ otherwise }
    \end{cases}
\end{equation}
with $\gamma = 2.0$ in our experiments, focusing on under-refined ground-truth and over-refined predicted areas.
\subsection{Temporal sparsity prediction}
\label{sec:sup_temp_spar}
A key aspect of our approach is the prediction of temporal sparsity to maintain consistency between frames. This module contrasts the feature maps of consecutive frames to predict their absolute differences. Comprising three convolution layers with batch normalization and ReLU activation, this module processes the concatenated feature maps from two adjacent frames and predicts the binary differences between them.

\begin{figure}[t]
    \centering
    \includegraphics[width=\columnwidth]{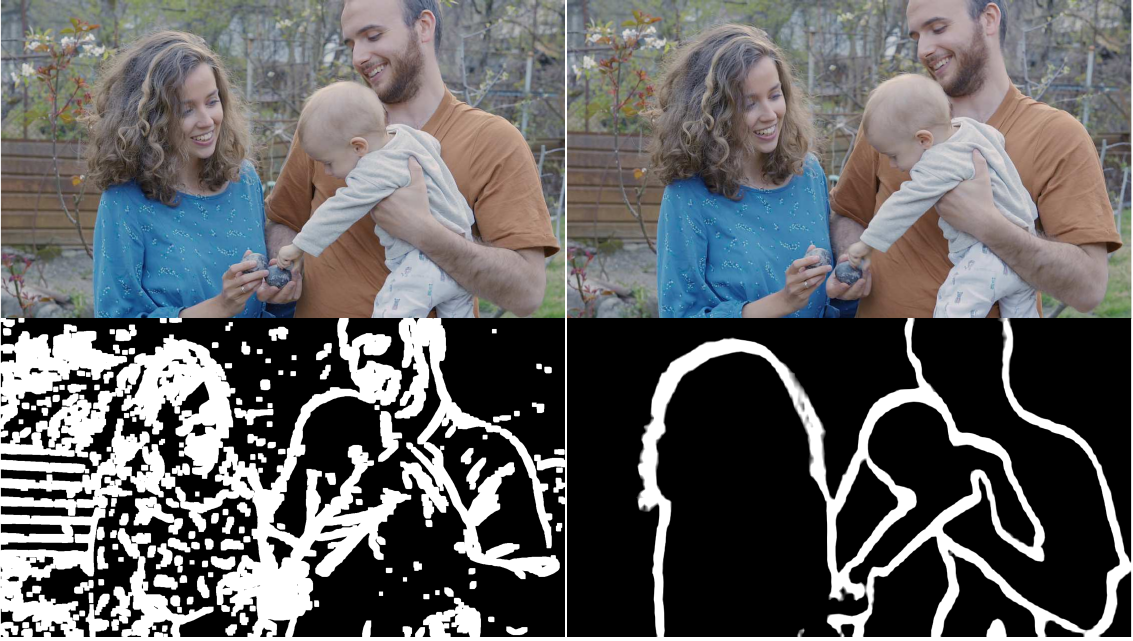}
    \begin{tblr}{width=\columnwidth,colspec={X[1,c]X[1,c]}}
        SparseMat~\cite{sun2023sparsemat} & Ours
    \end{tblr}
    \caption{\textbf{Temporal Sparsity Between Two Consecutive Frames}. The top row displays a pair of successive frames. Below, the second row illustrates the predicted differences by two distinct frameworks, with areas of discrepancy emphasized in white. In contrast to SparseMat's output, which appears cluttered and noisy, our module generates a more refined sparsity map. This map effectively accentuates the foreground regions that undergo notable changes between the frames, providing a clearer and more focused representation of temporal sparsity. (Best viewed in color).}
    \label{fig:sparsity}
    \vspace{-1em}
\end{figure}

Unlike SparseMat~\cite{sun2023sparsemat}, which relies on manual threshold selection for frame differences, our method offers a more robust and domain-independent approach to determining frame sparsity. This is particularly effective in handling variations in movement, resolution, and domain between frames, as demonstrated in~\Fref{fig:sparsity}

\subsection{Forward and backward matte fusion}
\label{sub:sup_forward_backward}
The forward-backward fusion for the $i$-th instance at frame $t$ is respectively:
\begin{equation}
    \begin{split}
        \mathbf{A}^f(t, i) &= \Delta(t) \x \mathbf{A}(t, i) \\
                        & + (1 - \Delta(t)) \x \mathbf{A}^f(t-1, i) \\
    \end{split}
\end{equation}
\begin{equation}
    \begin{split}\
        \mathbf{A}^b(t, i) &= \Delta(t+1) \x \mathbf{A}(t, i) \\
                        & + (1 - \Delta(t+1)) \x \mathbf{A}^b(t+1, i) \\
    \end{split}
\end{equation}
Each entry $j=(x,y,t,i)$ on final output $\mathbf{A}^{\text{temp}}$ is:
\begin{equation}
    \mathbf{A}^{\text{temp}}(j) = \begin{cases}
        \mathbf{A}(j), \text{ if } \mathbf{A}^f(j) \neq \mathbf{A}^b(j) \\
        \mathbf{A}^f(j), \text{ otherwise}
    \end{cases}
\end{equation}
This fusion enhances temporal consistency and minimizes error propagation.

\section{Image matting}

This section expands on the image matting process, providing additional insights into dataset generation and comprehensive comparisons with existing methods. We delve into the creation of I-HIM50K and M-HIM2K datasets, offer detailed quantitative analyses, and present further qualitative results to underscore the effectiveness of our approach.

\subsection{Dataset generation and preparation}
\label{sec:sup_image_data}

\begin{figure}[b]
    \centering
    \vspace{-1em}
    \includegraphics[width=\columnwidth]{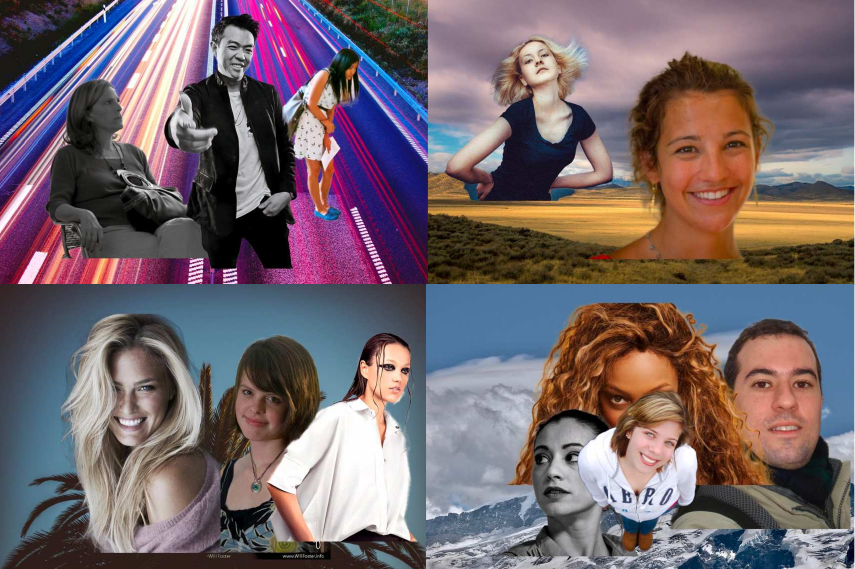}
    \caption{\textbf{Examples of I-HIM50K dataset}. (Best viewed in color).}
    \label{fig:sup_img_sample}
\end{figure}

The I-HIM50K dataset was synthesized from the HHM50K~\cite{sun2023sparsemat} dataset, which is known for its extensive collection of human image mattes. We employed a MaskRCNN~\cite{he2017mask} Resnet-50 FPN 3x model, trained on the COCO dataset, to filter out single-person images, resulting in a subset of 35,053 images. Following the InstMatt~\cite{sun2022instmatt} methodology, these images were composited against diverse backgrounds from the BG20K~\cite{li2022am2k} dataset, creating multi-instance scenarios with 2-5 subjects per image. The subjects were resized and positioned to maintain a realistic scale and avoid excessive overlap, as indicated by instance IoUs not exceeding 30\%. This process yielded 49,737 images, averaging 2.28 instances per image. During training, guidance masks were generated by binarizing the alpha mattes and applying random dropout, dilation, and erosion operations. Sample images from I-HIM50K are displayed in~\Fref{fig:sup_img_sample}. 

\begin{table}[t]
    \centering
    \caption{\textbf{Ten models with vary mask quality are used in M-HIM2K.} The MaskRCNN models are from detectron2 trained on COCO with different settings.}
    \begin{tblr}{width=\columnwidth,colspec={l|c}}
    \toprule
        Model &  COCO mask AP (\%)\\
        \hline
        r50\_c4\_3x & 34.4 \\
        r50\_dc5\_3x & 35.9 \\
        r101\_c4\_3x & 36.7 \\
        r50\_fpn\_3x & 37.2 \\
        r101\_fpn\_3x & 38.6 \\
        x101\_fpn\_3x & 39.5 \\
        r50\_fpn\_400e & 42.5 \\
        regnety\_400e & 43.3 \\
        regnetx\_400e & 43.5 \\
        r101\_fpn\_400e & 43.7 \\
    \bottomrule
    \end{tblr}    
    \label{tab:sup_image_data}
    \vspace{-1.5em}
\end{table}

The M-HIM2K dataset was designed to test model robustness against varying mask qualities. It comprises ten masks per instance, generated using various MaskRCNN models. More information about models used for this generation process is shown in~\Tref{tab:sup_image_data}. The masks were matched to instances based on the highest IoU with the ground truth alpha mattes, ensuring a minimum IoU threshold of 70\%. Masks that did not meet this threshold were artificially generated from ground truth. This process resulted in a comprehensive set of 134,240 masks, with 117,660 for composite and 16,600 for natural images, providing a robust benchmark for evaluating masked guided instance matting. The full dataset I-HIM50K and M-HIM2K will be released after the acceptance of this work.

\begin{table*}[t!]
    \centering
    \caption{\textbf{Full details of different input mask setting on HIM2K+M-HIM2K.} (Extension of~\Tref{tab:abl_mask_inp}). \textbf{Bold} denotes the lowest average error.}
    \scriptsize
    \begin{tblr}{width=\textwidth,colspec={@{}l|ccccccc|ccccccc|c@{}}}
    \toprule
         \SetCell[r=2]{c}{Mask input}
                & \SetCell[c=7]{c}{Composition} & & & & & & 
                    & \SetCell[c=7]{c}{Natural} & & & & & & 
                    & \SetCell[r=2]{c}{}\\
    \hline
    & MAD & MAD$_f$ & MAD$_u$ & MSE & SAD & Grad & Conn & MAD & MAD$_f$ & MAD$_u$ & MSE & SAD & Grad & Conn \\
    \hline
    \SetCell[r=2]{c}{Stacked} & 27.01 & 68.83 & 381.27 & 18.82 & 16.35 & 16.80 & 15.72 & 39.29 & 61.39 & 213.27 & 25.10 & 25.52 & 16.44 & 23.26 & mean \\ 
    & 0.83 & 5.93 & 7.06 & 0.76 & 0.50 & 0.31 & 0.51 & 4.21 & 13.37 & 14.10 & 4.01 & 2.00 & 0.70 & 2.02 & std \\ 
    \hline
    \SetCell[r=2]{c}{Embeded($C_e=1$)} & 19.18 & 68.04 & 330.06 & 12.40 & 11.64 & 13.00 & 11.16 & 33.60 & 60.35 & \textbf{188.44} & 20.63 & 21.40 & 13.44 & 19.18 & mean \\ 
    & 0.87 & 8.07 & 6.96 & 0.80 & 0.52 & 0.27 & 0.52 & 4.07 & 12.60 & 12.28 & 3.86 & 1.81 & 0.57 & 1.83 & std \\ 
    \hline
    \SetCell[r=2]{c}{Embeded($C_e=2$)} & 21.74 & 84.64 & 355.95 & 14.46 & 13.23 & 14.39 & 12.69 & 35.16 & 59.55 & 193.95 & 21.93 & 22.59 & 14.51 & 20.40 & mean \\ 
    & 0.92 & 7.33 & 7.68 & 0.85 & 0.55 & 0.27 & 0.55 & 4.23 & 13.79 & 13.45 & 4.03 & 2.31 & 0.61 & 2.32 & std \\ 
    \hline
    \SetCell[r=2]{c}{Embeded($C_e=3$)} & \textbf{17.75} & \textbf{53.23} & \textbf{315.43} & \textbf{11.19} &\textbf{10.79} & \textbf{12.52} & \textbf{10.32} & \textbf{33.06} & \textbf{56.69} & 189.59 & \textbf{20.22} & \textbf{19.43} & \textbf{13.11} & \textbf{17.30} & mean \\ 
    & 0.66 & 5.04 & 6.31 & 0.60 & 0.39 & 0.24 & 0.39 & 3.74 & 11.90 & 12.49 & 3.58 & 1.92 & 0.51 & 1.95 & std \\ 
    \hline
    \SetCell[r=2]{c}{Embeded($C_e=5$)} & 24.79 & 73.22 & 384.14 & 17.07 & 15.09 & 16.19 & 14.58 & 34.25 & 65.57 & 216.56 & 20.39 & 21.89 & 15.66 & 19.70 & mean \\ 
    & 0.88 & 4.99 & 7.24 & 0.79 & 0.52 & 0.30 & 0.52 & 4.16 & 13.59 & 13.09 & 3.96 & 2.31 & 0.58 & 2.32 & std \\
    \bottomrule
    \end{tblr}
    \label{tab:abl_mask_full}
\end{table*}

\begin{table*}[t!]
    \centering
    \caption{\textbf{Full details of different training objective components on HIM2K+M-HIM2K.} (Extension of~\Tref{tab:abl_loss}). \textbf{Bold} denotes the lowest average error.}
    \scriptsize
    \begin{tblr}{width=\textwidth,colspec={@{}cc|ccccccc|ccccccc|c@{}}}
    \toprule
         \SetCell[r=2]{c}{$\mathcal{L}_{att}$}
            & \SetCell[r=2]{c}{$\mathbf{W}_8$}
                & \SetCell[c=7]{c}{Composition} & & & & & & 
                    & \SetCell[c=7]{c}{Natural} & & & & & & 
                    & \SetCell[r=2]{c}{}\\
    \hline
    & & MAD & MAD$_f$ & MAD$_u$ & MSE & SAD & Grad & Conn & MAD & MAD$_f$ & MAD$_u$ & MSE & SAD & Grad & Conn \\
    \hline
    & & 31.77 & 52.70 & \textbf{294.22} & 24.13 & 18.92 & 16.58 & 18.27 & 46.68 & 51.23 & \textbf{176.60} & 33.61 & 32.89 & 15.68 & 30.64 & mean \\ 
    & & 0.90 & 4.92 & 5.24 & 0.85 & 0.54 & 0.26 & 0.54 & 3.64 & 10.27 & 9.58 & 3.47 & 1.85 & 0.50 & 1.85 & std \\ 
    \hline
    & \SetCell[r=2]{c}{\checkmark} & 25.41 & 104.24 & 342.19 & 18.36 & 15.29 & 14.53 & 14.75 & 46.30 & 87.18 & 210.72 & 32.93 & 31.40 & 15.84 & 29.26 & mean \\ 
    & & 0.72 & 6.15 & 5.53 & 0.67 & 0.43 & 0.23 & 0.43 & 3.71 & 11.68 & 10.62 & 3.55 & 1.85 & 0.50 & 1.86 & std \\ 
    \hline
    \SetCell[r=2]{c}{\checkmark} & & 17.56 & 53.51 & 302.07 & 11.24 & \textbf{10.65} & \textbf{12.34} & 10.22 & 32.95 & \textbf{51.11} & 183.13 & 20.41 & \textbf{19.23} & 13.29 & \textbf{17.06} & mean \\ 
    & & 0.75 & 6.32 & 6.32 & 0.70 & 0.45 & 0.27 & 0.45 & 3.34 & 10.25 & 10.99 & 3.19 & 2.04 & 0.60 & 2.06 & std \\ 
    \hline
    \SetCell[r=2]{c}{\checkmark}& \SetCell[r=2]{c}{\checkmark}& \textbf{17.55} & \textbf{47.81} & 301.96 & \textbf{11.23} & 10.68 & \textbf{12.34} & \textbf{10.19} & \textbf{32.03} & 53.15 & 183.42 & \textbf{19.42} & 19.60 & \textbf{13.16} & 17.43 & mean \\ 
    & & 0.68 & 5.21 & 5.73 & 0.63 & 0.41 & 0.25 & 0.41 & 3.48 & 10.77 & 11.18 & 3.32 & 1.92 & 0.55 & 1.94 & std \\
    \bottomrule
    \end{tblr}
    \label{tab:abl_loss_full}
\end{table*}

\begin{figure}[b]
    \centering
    \includegraphics[width=\columnwidth]{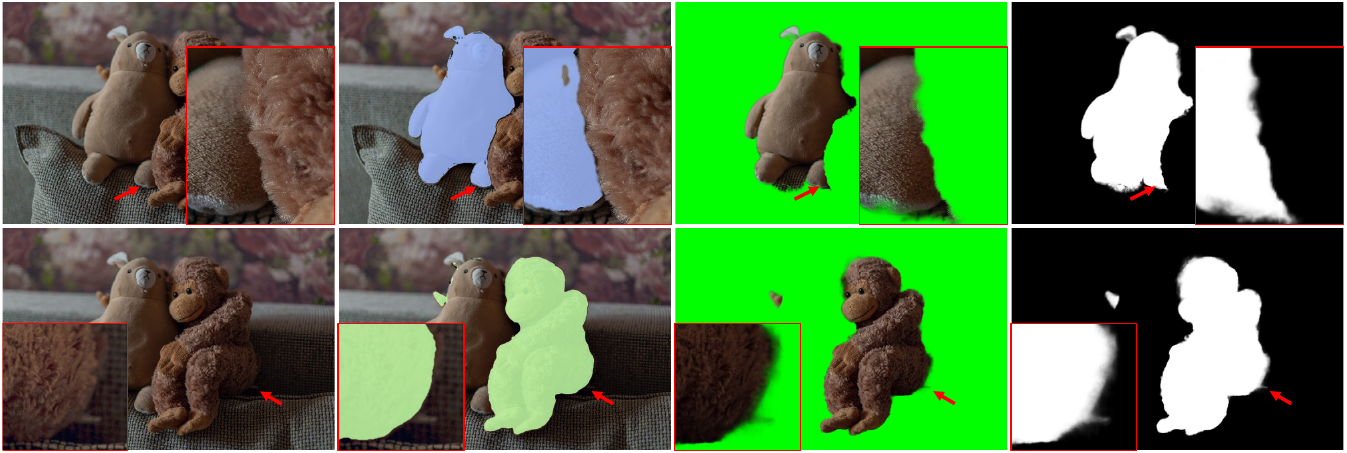}
    \includegraphics[width=\columnwidth]{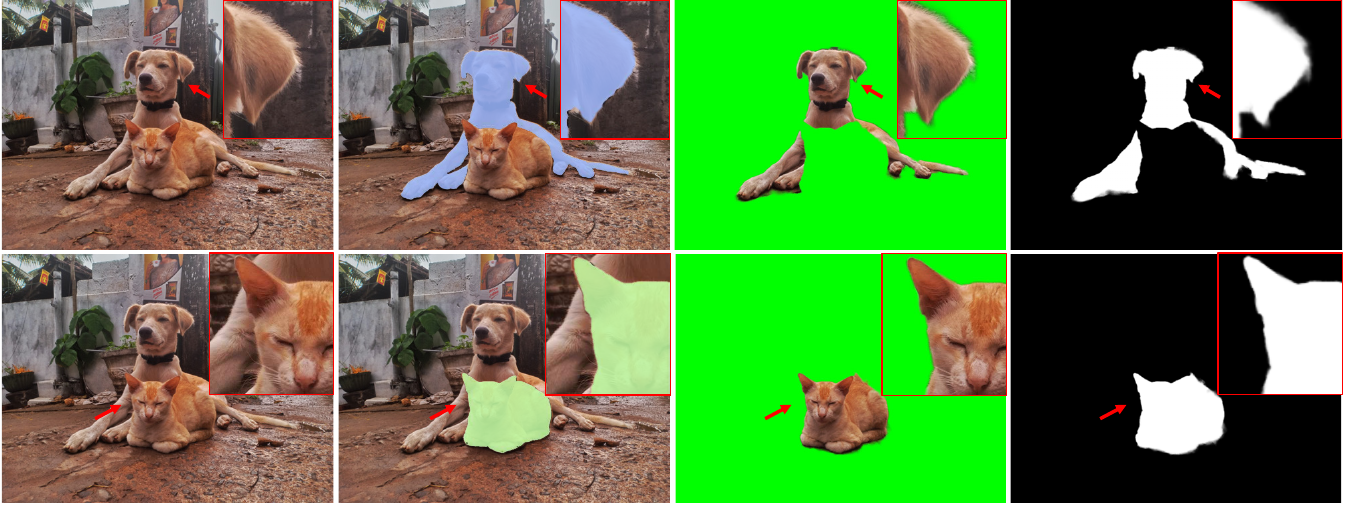}
    \begin{tblr}{width=\columnwidth,colspec={X[1,c]X[1,c]X[1,c]X[1,c]}}
        Image & Mask & Foreground & Alpha~matte
    \end{tblr}
    \caption{\textbf{Our framework can generalize to any object.} Without humans appearing in the image, our framework still performs the matting task very well to the mask-guided objects. (Best viewed in color and digital zoom).}
    \label{fig:sup_img_general2}
\end{figure}

\subsection{Training details}
\label{sec:sup_image_train_details}
We initialized our feature extractor with ImageNet~\cite{russakovsky2015imagenet} weights, following previous methods~\cite{yu2021mgm,sun2022instmatt}. Our models were retrained on the I-HIM50K dataset with a crop size 512. All baselines underwent 100 training epochs, using the HIM2K composition set for validation. The training was conducted on 4 A100 GPUs with a batch size 96. We employed AdamW for optimization, with a learning rate of $1.5 \times 10^{-4}$ and a cosine decay schedule post 1,500 warm-up iterations. The training also incorporated curriculum learning as MGM and standard augmentation as other baselines. During training, mask orders were shuffled, and some masks were randomly omitted. In testing, images were resized to have a short side of 576 pixels.

\subsection{Quantitative details}
\label{sec:sup_image_quan}

We extend the ablation study from the main paper, providing detailed statistics in~\Tref{tab:abl_mask_full} and~\Tref{tab:abl_loss_full}. These tables offer insights into the average and standard deviation of performance metrics across HIM2K~\cite{sun2022instmatt} and M-HIM2K datasets. Our model not only achieves competitive average results but also maintains low variability in performance across different error metrics. Additionally, we include the Sum Absolute Difference (SAD) metric, aligning with previous image matting benchmarks.

Comprehensive quantitative results comparing our model with baseline methods on HIM2K and M-HIM2K are presented in~\Tref{tab:details}. This analysis highlights the impact of mask quality on matting output, with our model demonstrating consistent performance even with varying mask inputs.

\begin{figure}[t!]
    \centering
    \includegraphics[width=\columnwidth]{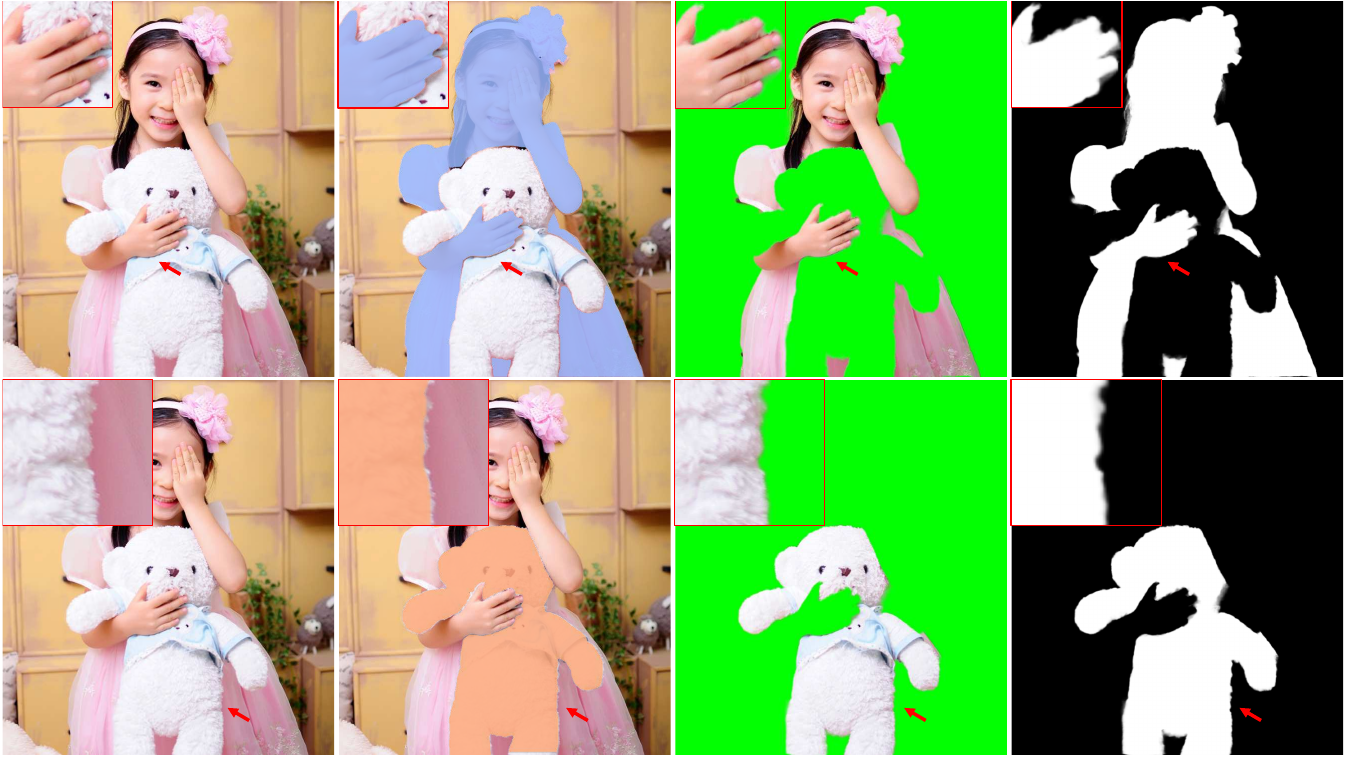}
    \includegraphics[width=\columnwidth]{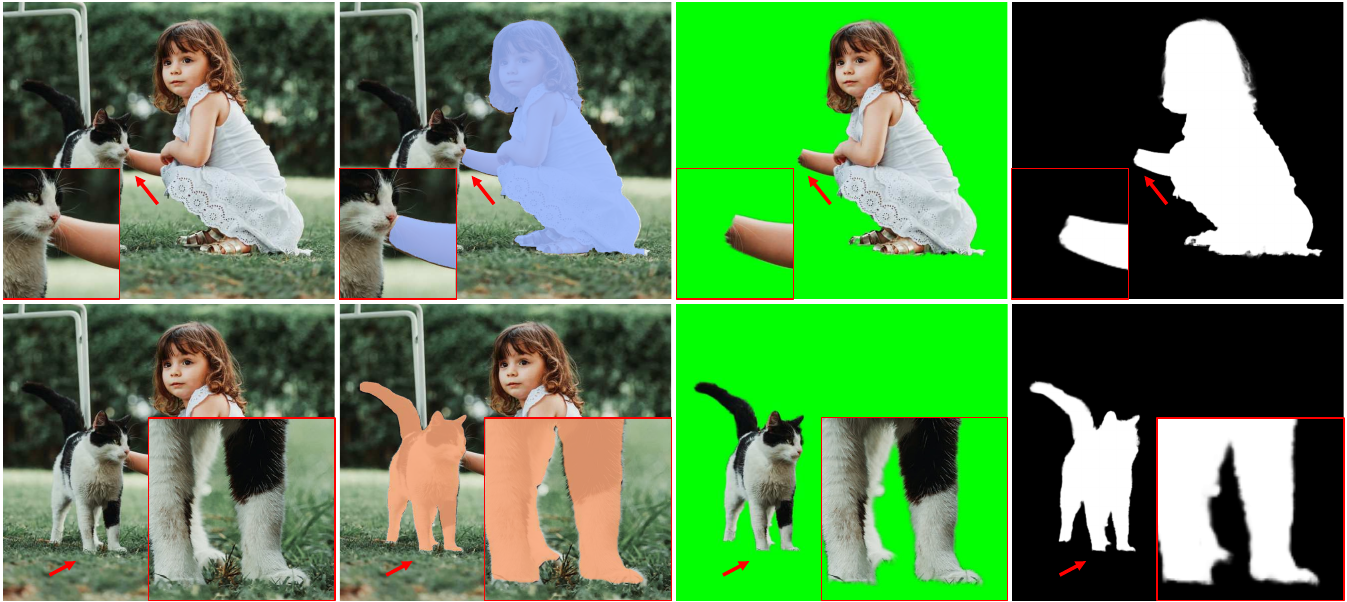}
    \includegraphics[width=\columnwidth]{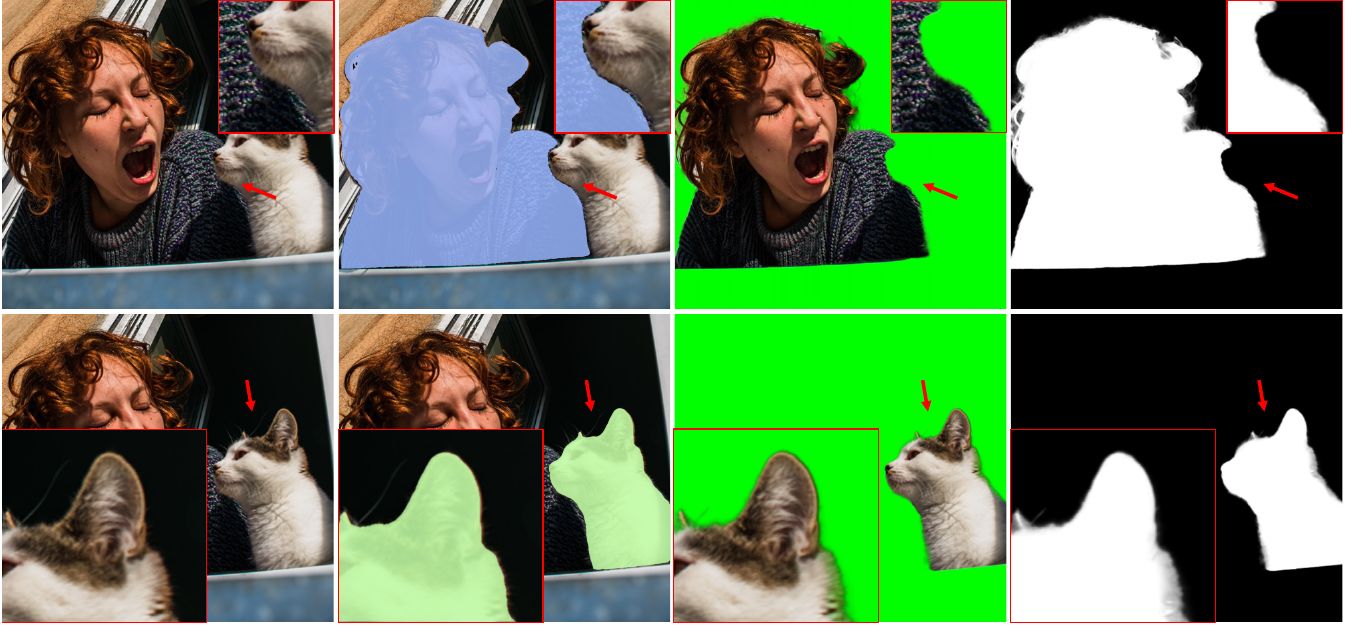}
    \begin{tblr}{width=\columnwidth,colspec={X[1,c]X[1,c]X[1,c]X[1,c]}}
        Image & Mask & Foreground & Alpha~matte
    \end{tblr}
    \caption{\textbf{Our solution is not limited to human instances.} When testing with other objects, our solution is able to produce fairly accurate alpha matte without training on them. (Best viewed in color and digital zoom).}
    \label{fig:sup_img_general1}
\end{figure}

\begin{table}[b]
    \centering
    \caption{\textbf{Compare between previous dense progressive refinement (PR) - MGM and our proposed guided sparse progressive refinement.} Numbers are mean on HIM2K+M-HIM2K and small numbers indicate the std.}
    \label{tab:abl_pr}
    \footnotesize 
    \begin{tblr}{width=\columnwidth,colsep=1.5pt,colspec={@{}c|cccccc}}
         \toprule
         PR & MAD & MSE & Grad & Conn & MAD$_f$ & MAD$_u$ \\
         \hline
         \SetCell[c=4]{l}{\textbf{\textit{Comp Set}}} \\
         \hline
         MGM & 14.70 \std{0.4} & 8.87 \std{0.3} & 10.39 \std{0.2} & 8.44 \std{0.2} & 32.02 \std{1.5} & 252.34 \std{4.2} \\
         Ours & \fst{12.93 \std{0.3}} & \fst{7.26 \std{0.3}} & \fst{8.91 \std{0.1}} & \fst{7.37 \std{0.2}} & \fst{19.54 \std{1.0}} & \fst{235.95 \std{3.4}} \\
         \hline
         \SetCell[c=4]{l}{\textbf{\textit{Natural Set}}} \\
         \hline
         MGM & 27.66 \std{4.1} & 16.94 \std{3.9} & 10.49 \std{0.7} & 13.95 \std{1.5} & 52.72 \std{12.1} & 150.71 \std{13.3}\\
         Ours & \fst{27.17 \std{3.3}} & \fst{16.09 \std{3.2}} & \fst{9.94 \std{0.6}} & \fst{13.42 \std{1.4}} & \fst{49.52 \std{8.0}} & \fst{146.71 \std{11.6}}\\
         \bottomrule
    \end{tblr}
\end{table}

We also perform another experiment when the MGM-style refinement replaces our proposed sparse guided progressive refinement. The~\Tref{tab:abl_pr} shows the results where our proposed method outperforms the previous approach in all metrics.

\subsection{More qualitative results on natural images}
\label{sec:sup_image_qual}

\begin{figure*}[t]
    \centering
    \includegraphics[width=\textwidth]{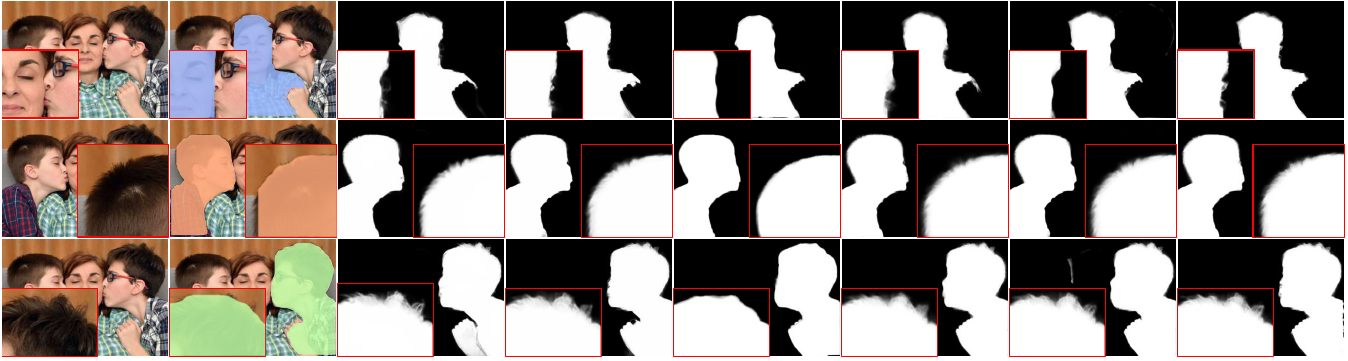}
    \includegraphics[width=\textwidth]{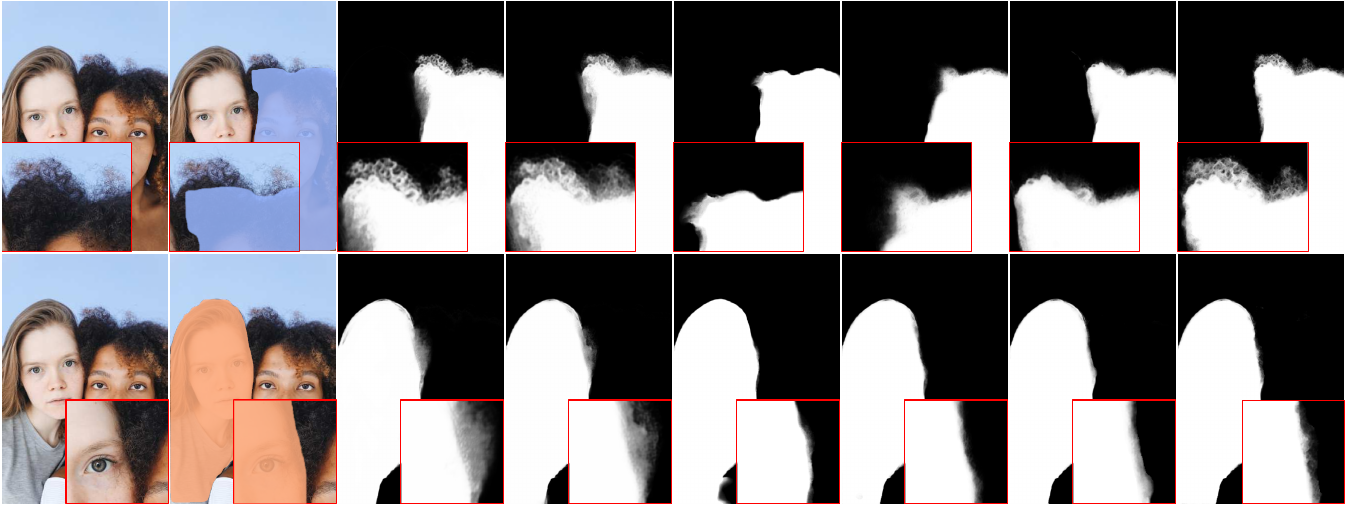}
    \includegraphics[width=\textwidth]{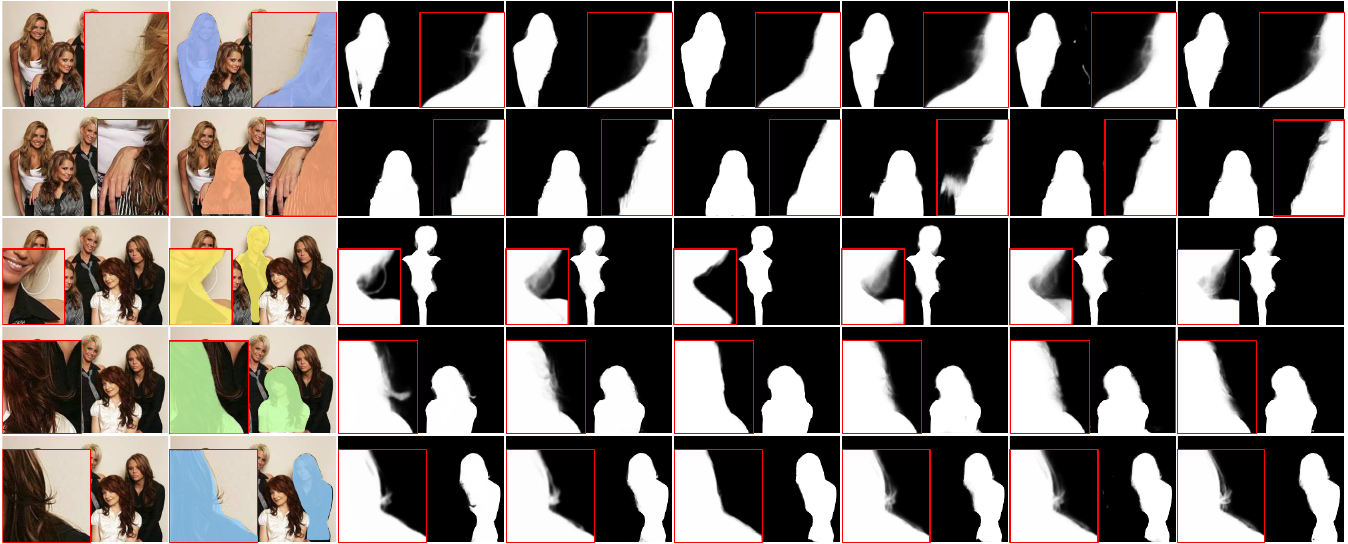}
    \begin{tblr}{width=\textwidth,colspec={X[1,c]X[1,c]X[1,c]X[1,c]X[1,c]X[1,c]X[1,c]X[1,c]}}
        Image & Mask & InstMatt~\cite{sun2022instmatt} (public) & InstMatt~\cite{sun2022instmatt} & {SparseMat \\ \cite{sun2023sparsemat}} & MGM~\cite{yu2021mgm} & MGM$^\star$ & Ours
    \end{tblr}
    \caption{\textbf{Our model produces highly detailed alpha matte on natural images.} Our results show that it is accurate and comparable with previous instance-agnostic and instance-awareness methods without expensive computational costs. {\color{red}Red} squares zoom in the detail regions for each instance. (Best viewed in color and digital zoom).}
    \label{fig:sup_img_detail}
\end{figure*}

\Fref{fig:sup_img_detail} showcases our model's performance in challenging scenarios, particularly in accurately rendering hair regions. Our framework consistently outperforms MGM$^\star$ in detail preservation, especially in complex instance interactions. In comparison with InstMatt, our model exhibits superior instance separation and detail accuracy in ambiguous regions.

\begin{figure*}[t]
    \centering
    \vspace{-2em}
    \includegraphics[width=\textwidth]{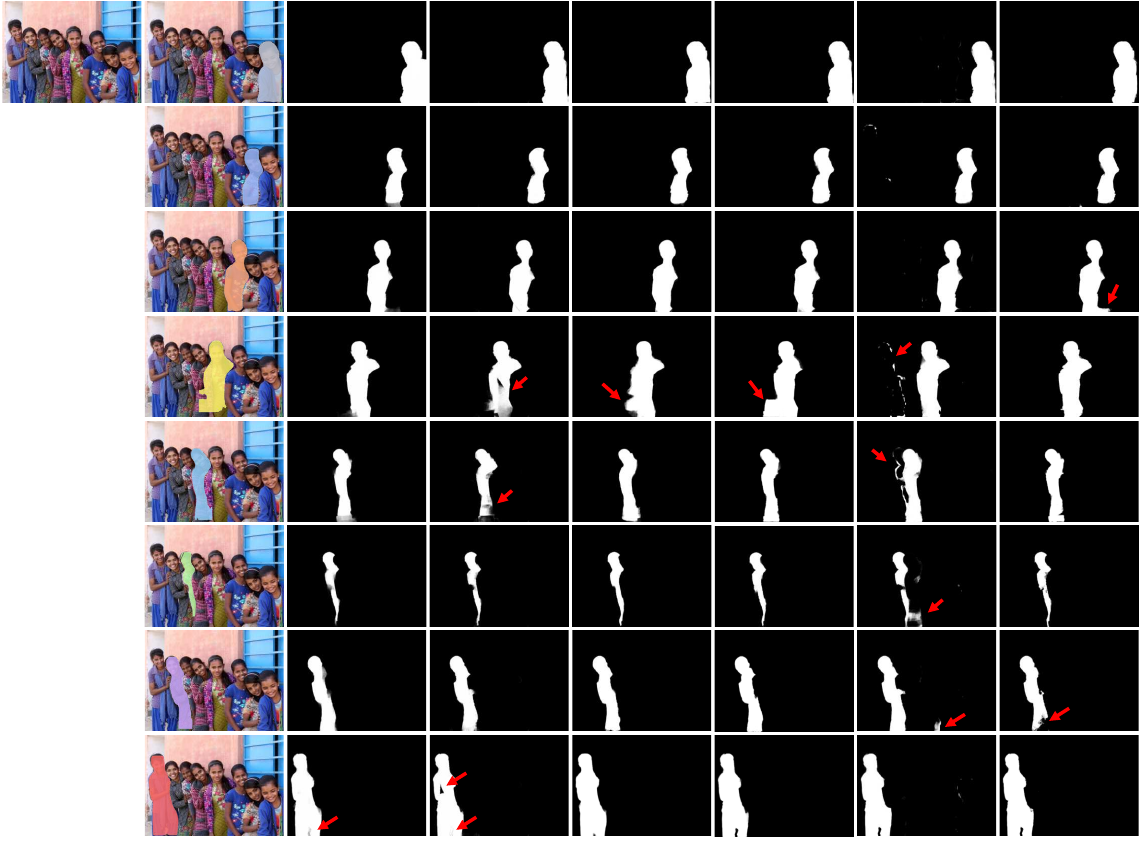}
    \begin{tblr}{width=\textwidth,colspec={X[1,c]X[1,c]X[1,c]X[1,c]X[1,c]X[1,c]X[1,c]X[1,c]}}
        Image & Mask & InstMatt~\cite{sun2022instmatt} (public) & InstMatt~\cite{sun2022instmatt} & {SparseMat \\ \cite{sun2023sparsemat}} & MGM~\cite{yu2021mgm} & MGM$^\star$ & Ours
    \end{tblr}
    \caption{\textbf{Our frameworks precisely separate instances in an extreme case with many instances.} While MGM often causes the overlapping between instances and MGM$^\star$ contains noises, ours produces on-par results with InstMatt trained on the external dataset. {\color{red} Red} arrow indicates the errors. (Best viewed in color and digital zoom).}
    \label{fig:sup_img_extreme1}
\end{figure*}

\begin{figure*}[t]
    \centering
    \vspace{-1.5em}
    \includegraphics[width=\textwidth]{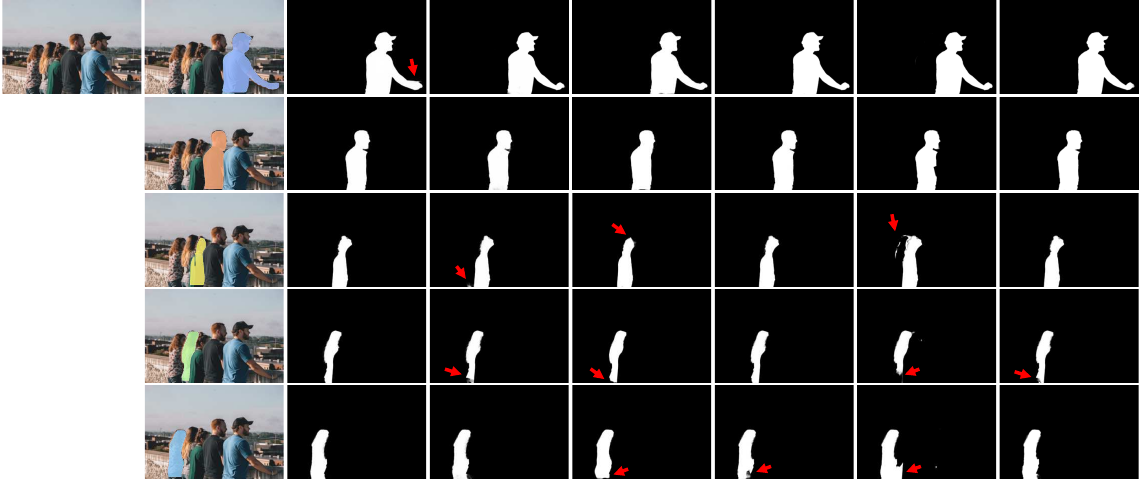}
    \begin{tblr}{width=\textwidth,colspec={X[1,c]X[1,c]X[1,c]X[1,c]X[1,c]X[1,c]X[1,c]X[1,c]}}
        Image & Mask & InstMatt~\cite{sun2022instmatt} (public) & InstMatt~\cite{sun2022instmatt} & {SparseMat \\ \cite{sun2023sparsemat}} & MGM~\cite{yu2021mgm} & MGM$^\star$ & Ours
    \end{tblr}
    \vspace{-1em}
    \caption{\textbf{Our frameworks precisely separate instances in a single pass.} The proposed solution shows comparable results with InstMatt and MGM without running the prediction/refinement five times. {\color{red} Red} arrow indicates the errors. (Best viewed in color and digital zoom).}
    \label{fig:sup_img_extreme2}
\end{figure*}

\Fref{fig:sup_img_extreme1} and \Fref{fig:sup_img_extreme2} illustrate the performance of our model and previous works in extreme cases involving multiple instances. While MGM$^\star$ struggles with noise and accuracy in dense instance scenarios, our model maintains high precision. InstMatt, without additional training data, shows limitations in these complex settings.

\begin{figure*}[t]
    \centering
    \includegraphics[width=\textwidth]{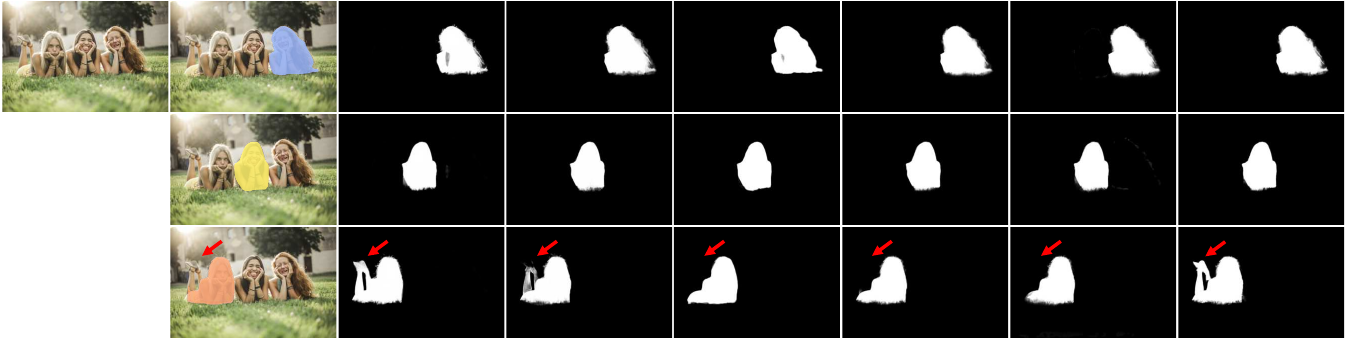}
    \begin{tblr}{width=\textwidth,colspec={X[1,c]X[1,c]X[1,c]X[1,c]X[1,c]X[1,c]X[1,c]X[1,c]}}
        Image & Mask & InstMatt~\cite{sun2022instmatt} (public) & InstMatt~\cite{sun2022instmatt} & {SparseMat \\ \cite{sun2023sparsemat}} & MGM~\cite{yu2021mgm} & MGM$^\star$ & Ours
    \end{tblr}
    \vspace{-1em}
    \caption{\textbf{Unlike MGM and SparseMat, our model is robust to the input guidance mask.} With the attention head, our model produces more stable results to mask inputs without complex refinement between instances like InstMatt. {\color{red} Red} arrow indicates the errors. (Best viewed in color and digital zoom).}
    \label{fig:sup_img_mask_robust}
\end{figure*}

\begin{figure*}[t]
    \centering
    \includegraphics[width=\textwidth]{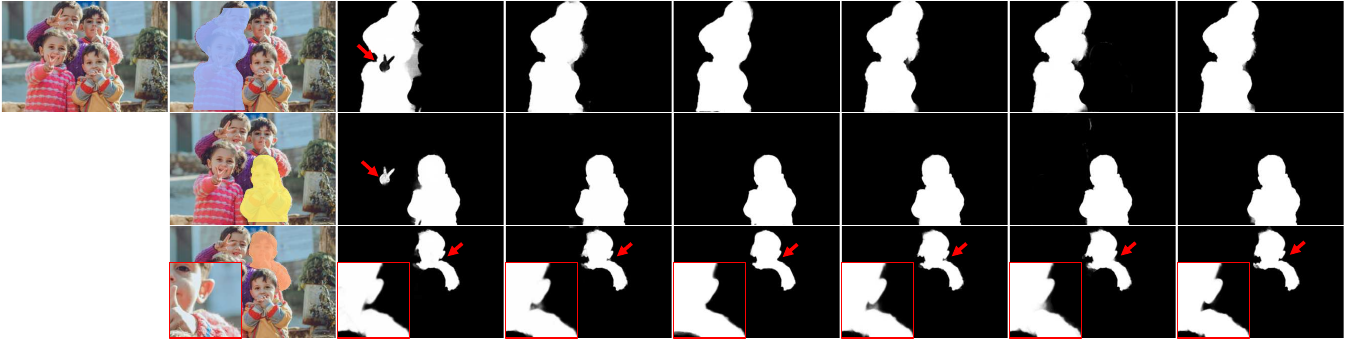}
    \begin{tblr}{width=\textwidth,colspec={X[1,c]X[1,c]X[1,c]X[1,c]X[1,c]X[1,c]X[1,c]X[1,c]}}
        Image & Mask & InstMatt~\cite{sun2022instmatt} (public) & InstMatt~\cite{sun2022instmatt} & {SparseMat \\ \cite{sun2023sparsemat}} & MGM~\cite{yu2021mgm} & MGM$^\star$ & Ours
    \end{tblr}
    \vspace{-1em}
    \caption{\textbf{Our solution works correctly with multi-instance mask guidances.} When multiple instances exist in one guidance mask, we still produce the correct union alpha matte for those instances. {\color{red} Red} arrow indicates the errors or the zoom-in region in {\color{red} red} box. (Best viewed in color and digital zoom).}
    \label{fig:sup_img_mask_multi}
    \vspace{-2em}
\end{figure*}

The robustness of our mask-guided approach is further demonstrated in~\Fref{fig:sup_img_mask_robust}. Here, we highlight the challenges faced by MGM variants and SparseMat in predicting missing parts in mask inputs, which our model addresses. However, it is important to note that our model is not designed as a human instance segmentation network. As shown in~\Fref{fig:sup_img_mask_multi}, our framework adheres to the input guidance, ensuring precise alpha matte prediction even with multiple instances in the same mask. 

Lastly, \Fref{fig:sup_img_general1} and \Fref{fig:sup_img_general2} emphasize our model's generalization capabilities. The model accurately extracts both human subjects and other objects from backgrounds, showcasing its versatility across various scenarios and object types.

All examples are Internet images without groundtruth and the mask from r101\_fpn\_400e are used as the guidance.

\onecolumn

\clearpage
\begin{longtblr}[
        theme=cvpr,
        caption={\textbf{Details of quantitative results on HIM2K+M-HIM2K} (Extension of~\Tref{tab:image_matting}). Gray indicates the public weight without retraining. },
        label={tab:details}
    ]{
        width=\textwidth,
        rows = {font=\scriptsize},
        row{4-15,41-52}={gainsboro},
        colspec={@{}c|ccccccc|ccccccc|l@{}}
    }
         \toprule
         \SetCell[r=2]{c}{Model} 
            & \SetCell[c=7]{c}{Composition set} & & & & & & 
            & \SetCell[c=7]{c}{Natural set} & & & & & &
            & \SetCell[r=2]{c}{Mask from} \\
         \hline
         & MAD & MAD$_f$ & MAD$_u$ & MSE & SAD & Grad & Conn & MAD & MAD$_f$ & MAD$_u$ & MSE & SAD & Grad & Conn \\
         \hline
         \SetCell[c=16]{l} \textbf{\textit{Instance-agnostic}} \\
         \hline
         \SetCell[r=12]{c}{MGM \\ \cite{park2023mgmwild}} & 25.79 & 69.67 & 331.73 & 17.00 & 15.65 & 13.64 & 14.91 & 48.05 & 103.81 & 233.85 & 32.66 & 27.44 & 14.72 & 25.07 & r50\_c4\_3x \\ 
            & 24.75 & 70.92 & 316.59 & 16.21 & 15.01 & 13.17 & 14.23 & 34.67 & 66.28 & 183.48 & 21.03 & 22.82 & 12.79 & 20.30 & r50\_dc5\_3x \\ 
            & 23.60 & 66.79 & 321.23 & 15.03 & 14.38 & 13.19 & 13.62 & 35.51 & 70.94 & 198.99 & 20.96 & 22.62 & 13.73 & 20.17 & r101\_c4\_3x \\ 
            & 24.55 & 67.27 & 316.29 & 15.97 & 14.91 & 13.14 & 14.12 & 33.66 & 67.41 & 184.99 & 19.93 & 21.99 & 13.06 & 19.43 & r50\_fpn\_3x \\ 
            & 23.42 & 66.37 & 310.99 & 14.94 & 14.21 & 12.84 & 13.42 & 35.14 & 72.30 & 183.87 & 21.02 & 21.87 & 12.82 & 19.34 & r101\_fpn\_3x \\ 
            & 22.71 & 63.35 & 305.67 & 14.36 & 13.81 & 12.64 & 13.03 & 31.06 & 61.76 & 175.33 & 17.60 & 20.98 & 12.61 & 18.44 & x101\_fpn\_3x \\ 
            & 22.03 & 61.91 & 300.29 & 13.85 & 13.36 & 12.30 & 12.59 & 29.16 & 57.59 & 165.22 & 15.93 & 20.10 & 11.76 & 17.56 & r50\_fpn\_400e \\ 
            & 21.37 & 57.28 & 296.73 & 13.18 & 12.98 & 12.16 & 12.21 & 26.40 & 51.24 & 158.95 & 13.42 & 17.73 & 11.45 & 15.10 & regnety\_400e \\ 
            & 21.78 & 60.31 & 297.14 & 13.62 & 13.22 & 12.25 & 12.46 & 27.09 & 49.26 & 160.05 & 13.82 & 17.48 & 11.20 & 14.87 & regnetx\_400e \\ 
            & 21.52 & 60.07 & 297.14 & 13.44 & 13.14 & 12.20 & 12.38 & 24.41 & 51.46 & 152.90 & 11.62 & 17.43 & 11.09 & 14.84 & r101\_fpn\_400e \\ 
            \hline
            & 23.15 & 64.39 & 309.38 & 14.76 & 14.07 & 12.75 & 13.30 & 32.52 & 65.20 & 179.76 & 18.80 & 21.05 & 12.52 & 18.51 & \textbf{mean} \\ 
            & 1.52 & 4.49 & 12.01 & 1.30 & 0.92 & 0.52 & 0.92 & 6.74 & 15.94 & 23.87 & 5.99 & 3.09 & 1.17 & 3.16 & \textbf{std} \\ 
            \bottomrule
            \SetCell[r=12]{c}{MGM \\ \cite{yu2021mgm}} & 15.94 & 32.55 & 266.64 & 9.62 & 9.68 & 10.11 & 9.18 & 37.55 & 86.64 & 191.09 & 24.03 & 21.15 & 11.34 & 18.94 & r50\_c4\_3x \\ 
            & 16.05 & 36.36 & 264.96 & 9.81 & 9.75 & 10.10 & 9.26 & 32.58 & 68.52 & 172.83 & 19.58 & 20.17 & 10.92 & 17.80 & r50\_dc5\_3x \\ 
            & 15.40 & 30.89 & 264.28 & 9.17 & 9.37 & 10.01 & 8.90 & 31.24 & 69.59 & 175.67 & 18.15 & 18.57 & 10.83 & 16.26 & r101\_c4\_3x \\ 
            & 15.93 & 34.54 & 265.44 & 9.68 & 9.67 & 10.10 & 9.20 & 32.83 & 75.06 & 173.63 & 19.72 & 19.13 & 10.85 & 16.81 & r50\_fpn\_3x \\ 
            & 15.74 & 34.23 & 263.35 & 9.50 & 9.55 & 10.02 & 9.07 & 30.77 & 69.10 & 171.92 & 17.78 & 18.22 & 10.67 & 15.95 & r101\_fpn\_3x \\ 
            & 15.23 & 36.18 & 260.80 & 9.03 & 9.27 & 9.92 & 8.76 & 30.09 & 63.23 & 167.58 & 17.34 & 18.51 & 10.69 & 16.09 & x101\_fpn\_3x \\ 
            & 14.96 & 34.13 & 259.17 & 8.81 & 9.08 & 9.83 & 8.61 & 28.28 & 50.35 & 158.02 & 15.71 & 17.71 & 10.24 & 15.25 & r50\_fpn\_400e \\ 
            & 14.53 & 31.71 & 256.33 & 8.41 & 8.83 & 9.73 & 8.35 & 26.95 & 49.55 & 155.63 & 14.43 & 15.69 & 9.98 & 13.34 & regnety\_400e \\ 
            & 14.82 & 33.06 & 257.09 & 8.69 & 9.01 & 9.80 & 8.53 & 26.61 & 47.81 & 154.05 & 14.22 & 15.45 & 9.87 & 13.16 & regnetx\_400e \\ 
            & 14.65 & 31.71 & 256.29 & 8.53 & 8.94 & 9.74 & 8.46 & 25.42 & 51.73 & 153.11 & 13.03 & 15.73 & 9.90 & 13.44 & r101\_fpn\_400e \\
            \hline
            & 15.32 & 33.54 & 261.43 & 9.13 & 9.31 & 9.94 & 8.83 & 30.23 & 63.16 & 167.35 & 17.40 & 18.03 & 10.53 & 15.70 & \textbf{mean} \\ 
            & 0.57 & 1.88 & 4.00 & 0.51 & 0.34 & 0.15 & 0.34 & 3.62 & 12.97 & 12.14 & 3.26 & 1.93 & 0.50 & 1.94 & \textbf{std} \\ 
            \bottomrule
            \pagebreak
            \SetCell[r=12]{c}{SparseMat \\ \cite{sun2023sparsemat}} & 23.14 & 47.59 & 378.89 & 16.37 & 13.97 & 15.56 & 13.54 & 46.28 & 101.48 & 255.98 & 31.99 & 26.81 & 17.97 & 24.82 & r50\_c4\_3x \\ 
            & 21.94 & 49.48 & 358.08 & 15.36 & 13.24 & 14.90 & 12.80 & 36.93 & 67.62 & 213.46 & 23.76 & 22.11 & 16.05 & 20.01 & r50\_dc5\_3x \\ 
            & 21.78 & 43.36 & 368.59 & 15.15 & 13.16 & 15.21 & 12.72 & 38.32 & 77.98 & 234.69 & 24.51 & 22.83 & 17.19 & 20.78 & r101\_c4\_3x \\ 
            & 21.94 & 47.00 & 361.30 & 15.33 & 13.24 & 14.99 & 12.80 & 37.16 & 74.18 & 218.62 & 23.95 & 21.95 & 16.39 & 19.86 & r50\_fpn\_3x \\ 
            & 21.43 & 46.51 & 356.43 & 14.88 & 12.93 & 14.81 & 12.48 & 35.95 & 72.78 & 218.46 & 22.62 & 20.67 & 16.11 & 18.58 & r101\_fpn\_3x \\ 
            & 20.63 & 47.73 & 349.81 & 14.12 & 12.48 & 14.58 & 12.02 & 34.32 & 64.51 & 209.64 & 21.10 & 20.44 & 16.03 & 18.33 & x101\_fpn\_3x \\ 
            & 20.29 & 44.20 & 342.14 & 13.93 & 12.22 & 14.21 & 11.76 & 31.44 & 57.51 & 197.53 & 18.58 & 19.49 & 14.96 & 17.35 & r50\_fpn\_400e \\ 
            & 19.65 & 41.20 & 340.38 & 13.29 & 11.85 & 14.08 & 11.38 & 30.21 & 48.53 & 194.90 & 17.32 & 17.47 & 14.82 & 15.31 & regnety\_400e \\ 
            & 19.90 & 41.40 & 336.40 & 13.56 & 12.02 & 14.03 & 11.56 & 29.85 & 52.17 & 191.09 & 16.99 & 17.19 & 14.52 & 15.03 & regnetx\_400e \\ 
            & 19.81 & 43.43 & 337.43 & 13.50 & 12.01 & 14.05 & 11.55 & 29.83 & 61.40 & 191.89 & 17.07 & 17.13 & 14.48 & 14.96 & r101\_fpn\_400e \\ 
            \hline
            & 21.05 & 45.19 & 352.95 & 14.55 & 12.71 & 14.64 & 12.26 & 35.03 & 67.82 & 212.63 & 21.79 & 20.61 & 15.85 & 18.50 & \textbf{mean} \\ 
            & 1.17 & 2.85 & 14.24 & 1.02 & 0.70 & 0.54 & 0.71 & 5.13 & 15.19 & 20.77 & 4.68 & 3.03 & 1.16 & 3.08 & \textbf{std} \\ 
            \bottomrule
            \SetCell[c=16]{l} \textbf{\textit{Instance-awareness}} \\
            \hline
            \SetCell[r=12]{c}{InstMatt \\ \cite{sun2022instmatt}} & 12.98 & 23.71 & 257.74 & 5.76 & 7.94 & 9.47 & 7.27 & 31.15 & 60.03 & 174.10 & 15.91 & 18.12 & 10.64 & 15.73 & r50\_c4\_3x \\ 
            & 13.15 & 23.08 & 257.38 & 5.96 & 8.05 & 9.48 & 7.38 & 28.05 & 51.53 & 164.19 & 13.63 & 16.89 & 10.33 & 14.53 & r50\_dc5\_3x \\ 
            & 12.99 & 22.42 & 257.52 & 5.79 & 7.93 & 9.47 & 7.26 & 27.06 & 48.52 & 162.72 & 12.90 & 16.06 & 10.29 & 13.68 & r101\_c4\_3x \\ 
            & 13.13 & 20.60 & 256.70 & 5.90 & 8.03 & 9.47 & 7.36 & 28.31 & 49.87 & 164.16 & 13.97 & 16.86 & 10.37 & 14.49 & r50\_fpn\_3x \\ 
            & 13.04 & 23.98 & 257.51 & 5.85 & 7.96 & 9.45 & 7.28 & 28.92 & 59.32 & 168.72 & 14.37 & 16.98 & 10.40 & 14.64 & r101\_fpn\_3x \\ 
            & 12.77 & 22.16 & 255.33 & 5.63 & 7.83 & 9.40 & 7.16 & 27.02 & 46.39 & 162.89 & 12.82 & 16.49 & 10.27 & 14.08 & x101\_fpn\_3x \\ 
            & 12.61 & 21.31 & 254.27 & 5.55 & 7.71 & 9.36 & 7.05 & 25.33 & 44.84 & 157.03 & 11.23 & 15.54 & 9.97 & 13.18 & r50\_fpn\_400e \\ 
            & 12.58 & 23.53 & 253.85 & 5.57 & 7.69 & 9.35 & 7.03 & 24.34 & 41.62 & 154.89 & 10.65 & 15.22 & 10.00 & 12.85 & regnety\_400e \\ 
            & 12.59 & 20.48 & 252.68 & 5.53 & 7.71 & 9.35 & 7.04 & 24.18 & 40.96 & 154.69 & 10.09 & 14.68 & 9.82 & 12.28 & regnetx\_400e \\ 
            & 12.67 & 21.14 & 253.13 & 5.60 & 7.75 & 9.35 & 7.09 & 23.22 & 43.23 & 151.78 & 9.67 & 15.00 & 9.88 & 12.60 & r101\_fpn\_400e \\ 
            \hline
            & 12.85 & 22.24 & 255.61 & 5.71 & 7.86 & 9.41 & 7.19 & 26.76 & 48.63 & 161.52 & 12.52 & 16.18 & 10.20 & 13.81 & \textbf{mean} \\ 
            & 0.23 & 1.31 & 2.00 & 0.16 & 0.14 & 0.06 & 0.13 & 2.48 & 6.76 & 6.94 & 2.05 & 1.08 & 0.26 & 1.08 & \textbf{std} \\ 
            \bottomrule
            \SetCell[r=12]{c}{InstMatt \\ \cite{sun2022instmatt}} & 18.23 & 57.23 & 298.66 & 10.51 & 11.06 & 11.33 & 10.45 & 37.91 & 86.84 & 202.20 & 22.28 & 21.31 & 12.22 & 19.11 & r50\_c4\_3x \\ 
            & 17.85 & 58.98 & 291.50 & 10.38 & 10.87 & 11.13 & 10.27 & 30.10 & 63.83 & 173.94 & 15.90 & 18.01 & 11.25 & 15.82 & r50\_dc5\_3x \\ 
            & 17.25 & 51.21 & 292.66 & 9.80 & 10.50 & 11.13 & 9.90 & 30.22 & 59.65 & 178.94 & 15.62 & 17.49 & 11.55 & 15.23 & r101\_c4\_3x \\ 
            & 17.69 & 55.80 & 292.90 & 10.22 & 10.80 & 11.19 & 10.19 & 30.27 & 60.16 & 175.66 & 16.44 & 17.38 & 11.33 & 15.13 & r50\_fpn\_3x \\ 
            & 17.18 & 55.67 & 288.95 & 9.85 & 10.45 & 11.02 & 9.84 & 28.80 & 60.88 & 170.89 & 14.55 & 16.88 & 11.12 & 14.69 & r101\_fpn\_3x \\ 
            & 16.65 & 53.37 & 284.66 & 9.41 & 10.16 & 10.85 & 9.56 & 27.77 & 55.06 & 168.20 & 14.14 & 16.91 & 11.04 & 14.70 & x101\_fpn\_3x \\ 
            & 16.29 & 52.00 & 281.15 & 9.21 & 9.88 & 10.69 & 9.29 & 25.51 & 52.89 & 156.40 & 12.15 & 15.90 & 10.47 & 13.70 & r50\_fpn\_400e \\ 
            & 15.99 & 50.92 & 279.15 & 8.97 & 9.71 & 10.65 & 9.12 & 24.82 & 45.83 & 156.46 & 11.83 & 15.14 & 10.43 & 12.94 & regnety\_400e \\ 
            & 16.47 & 51.85 & 280.00 & 9.37 & 10.01 & 10.69 & 9.42 & 23.73 & 47.85 & 153.70 & 10.35 & 14.69 & 10.17 & 12.49 & regnetx\_400e \\ 
            & 16.30 & 50.58 & 279.40 & 9.29 & 9.95 & 10.63 & 9.36 & 22.47 & 45.33 & 150.96 & 9.72 & 14.71 & 10.17 & 12.50 & r101\_fpn\_400e \\ 
            \hline
            & 16.99 & 53.76 & 286.90 & 9.70 & 10.34 & 10.93 & 9.74 & 28.16 & 57.83 & 168.74 & 14.30 & 16.84 & 10.98 & 14.63 & \textbf{mean} \\ 
            & 0.76 & 2.96 & 6.95 & 0.53 & 0.47 & 0.26 & 0.46 & 4.45 & 12.15 & 15.45 & 3.65 & 1.97 & 0.66 & 1.97 & \textbf{std} \\ 
            \bottomrule
            \SetCell[r=12]{c}{MGM$^\star$} & 14.87 & 46.70 & 256.01 & 8.32 & 8.99 & 10.31 & 8.32 & 37.36 & 65.40 & 181.68 & 23.97 & 20.50 & 11.66 & 17.45 & r50\_c4\_3x \\ 
            & 14.65 & 43.00 & 253.75 & 8.21 & 8.87 & 10.25 & 8.22 & 33.70 & 60.48 & 172.03 & 20.83 & 18.51 & 11.29 & 15.93 & r50\_dc5\_3x \\ 
            & 14.36 & 38.88 & 252.30 & 7.89 & 8.71 & 10.19 & 8.04 & 33.95 & 60.54 & 173.47 & 20.59 & 17.94 & 11.24 & 15.30 & r101\_c4\_3x \\ 
            & 14.68 & 44.85 & 254.50 & 8.21 & 8.88 & 10.24 & 8.22 & 33.29 & 54.82 & 170.89 & 20.21 & 18.28 & 11.27 & 15.55 & r50\_fpn\_3x \\ 
            & 14.70 & 44.68 & 254.29 & 8.24 & 8.89 & 10.21 & 8.25 & 32.07 & 68.47 & 171.41 & 18.80 & 17.44 & 11.07 & 14.84 & r101\_fpn\_3x \\ 
            & 14.27 & 43.56 & 251.19 & 7.83 & 8.68 & 10.13 & 8.00 & 30.96 & 50.90 & 166.14 & 18.02 & 17.53 & 11.07 & 14.91 & x101\_fpn\_3x \\ 
            & 13.94 & 38.70 & 248.02 & 7.58 & 8.46 & 10.00 & 7.79 & 29.86 & 48.23 & 158.22 & 16.92 & 16.91 & 10.79 & 14.32 & r50\_fpn\_400e \\ 
            & 13.57 & 39.12 & 246.18 & 7.24 & 8.21 & 9.89 & 7.56 & 28.53 & 46.70 & 156.07 & 15.84 & 15.98 & 10.52 & 13.38 & regnety\_400e \\ 
            & 14.11 & 41.69 & 247.92 & 7.75 & 8.57 & 10.00 & 7.91 & 27.17 & 41.88 & 150.59 & 14.42 & 15.35 & 10.36 & 12.75 & regnetx\_400e \\ 
            & 13.95 & 38.26 & 246.60 & 7.60 & 8.48 & 9.95 & 7.83 & 26.89 & 41.53 & 150.85 & 14.23 & 15.74 & 10.42 & 13.12 & r101\_fpn\_400e \\
            \hline
            & 14.31 & 41.94 & 251.08 & 7.89 & 8.67 & 10.12 & 8.01 & 31.38 & 53.89 & 165.13 & 18.38 & 17.42 & 10.97 & 14.75 & \textbf{mean} \\ 
            & 0.42 & 3.05 & 3.63 & 0.35 & 0.24 & 0.15 & 0.24 & 3.34 & 9.56 & 10.59 & 3.11 & 1.53 & 0.43 & 1.43 & \textbf{std} \\ 
            \bottomrule
            \SetCell[r=12]{c}{Ours} & 13.13 & 17.81 & 239.98 & 7.41 & 7.92 & 9.05 & 7.47 & 34.54 & 64.64 & 171.51 & 23.05 & 18.36 & 11.02 & 16.23 & r50\_c4\_3x \\ 
            & 13.28 & 21.29 & 238.15 & 7.61 & 8.03 & 9.03 & 7.58 & 27.66 & 52.90 & 149.52 & 16.56 & 16.05 & 10.15 & 13.90 & r50\_dc5\_3x \\ 
            & 13.20 & 19.24 & 240.33 & 7.49 & 7.98 & 9.07 & 7.53 & 29.04 & 54.52 & 154.34 & 17.75 & 16.72 & 10.53 & 14.58 & r101\_c4\_3x \\ 
            & 13.20 & 19.37 & 237.53 & 7.52 & 7.98 & 8.98 & 7.53 & 28.50 & 53.64 & 150.67 & 17.37 & 15.91 & 10.18 & 13.74 & r50\_fpn\_3x \\ 
            & 13.02 & 20.89 & 238.27 & 7.35 & 7.91 & 8.98 & 7.45 & 28.32 & 52.55 & 150.76 & 17.21 & 15.87 & 10.12 & 13.71 & r101\_fpn\_3x \\ 
            & 12.98 & 19.27 & 236.44 & 7.32 & 7.87 & 8.93 & 7.41 & 27.12 & 51.27 & 146.81 & 16.12 & 15.92 & 10.00 & 13.76 & x101\_fpn\_3x \\ 
            & 12.65 & 19.92 & 233.05 & 7.01 & 7.64 & 8.80 & 7.18 & 24.72 & 44.25 & 137.65 & 13.83 & 14.83 & 9.60 & 12.68 & r50\_fpn\_400e \\ 
            & 12.55 & 19.59 & 231.94 & 6.93 & 7.58 & 8.73 & 7.12 & 24.99 & 41.32 & 139.09 & 14.02 & 14.32 & 9.38 & 12.15 & regnety\_400e \\ 
            & 12.60 & 19.04 & 231.50 & 6.96 & 7.65 & 8.78 & 7.19 & 23.64 & 39.60 & 134.20 & 12.69 & 14.12 & 9.27 & 11.94 & regnetx\_400e \\ 
            & 12.69 & 19.01 & 232.26 & 7.05 & 7.69 & 8.78 & 7.23 & 23.16 & 40.47 & 132.55 & 12.25 & 13.67 & 9.17 & 11.49 & r101\_fpn\_400e \\ 
            \hline
            & 12.93 & 19.54 & 235.95 & 7.26 & 7.82 & 8.91 & 7.37 & 27.17 & 49.52 & 146.71 & 16.09 & 15.58 & 9.94 & 13.42 & \textbf{mean} \\ 
            & 0.28 & 0.99 & 3.44 & 0.25 & 0.17 & 0.13 & 0.17 & 3.34 & 7.95 & 11.60 & 3.16 & 1.39 & 0.59 & 1.41 & \textbf{std} \\
         \bottomrule
\end{longtblr}

\twocolumn

\begin{table}[t]
    \centering
    \scriptsize
    \caption{\textbf{The effectiveness of proposed temporal consistency modules on V-HIM60} (Extension of~\Tref{tab:abl_temp}). The combination of bi-directional Conv-GRU and forward-backward fusion achieves the best overall performance on three test sets. \textbf{Bold} highlights the best for each level.}
    \begin{tblr}{width=\columnwidth,colsep=1.2pt,colspec={@{}cc|cc|ccccccccc@{}}}
        \toprule
         \SetCell[c=2]{c}{Conv-GRU} & 
        & \SetCell[c=2]{c}{Fusion} & 
        &  \SetCell[r=2]{c}{MAD} & \SetCell[r=2]{c}{MAD$_f$} & \SetCell[r=2]{c}{MAD$_u$} & \SetCell[r=2]{c}{MSE} & \SetCell[r=2]{c}{SAD} & \SetCell[r=2]{c}{Grad} & \SetCell[r=2]{c}{Conn} & \SetCell[r=2]{c}{dtSSD} & \SetCell[r=2]{c}{MESSDdt} \\
        \hline
        Single & Bi & $\mathbf{\hat{A}}^f$ &  $\mathbf{\hat{A}}^b$ \\
        \hline
        \SetCell[c=9]{l}{\textbf{\textit{Easy level}}} \\
        \hline
        &  &  &  & 10.26 & 13.64 & 192.97 & 4.08 & 3.73 & 4.12 & 3.47 & 16.57 & 16.55 \\ 
        \checkmark &  &  &  & 10.15 & 12.83 & 192.69 & 4.03 & 3.71 & 4.09 & 3.44 & 16.42 & 16.44 \\ 
        & \checkmark &  &  & 10.14 & 12.70 & 192.67 & 4.05 & 3.70 & 4.09 & 3.44 & 16.41 & 16.42 \\ 
        & \checkmark & \checkmark &  & 11.32 & 20.13 & 194.27 & 5.01 & 4.10 & 4.67 & 3.85 & 16.51 & 17.85 \\ 
        & \checkmark & \checkmark & \checkmark & \fst{10.12} & \fst{12.60} & \fst{192.63} & \fst{4.02} & \fst{3.68} & \fst{4.08} & \fst{3.43} & \fst{16.40} & \fst{16.41} \\ 
        \bottomrule
        \SetCell[c=9]{l}{\textbf{\textit{Medium level}}} \\
        \hline
        &  &  &  & 13.88 & 4.78 & 202.20 & 5.27 & 5.56 & 6.30 & \fst{5.11} & 23.67 & 38.90 \\ 
        \checkmark &  &  &  & 13.84 & 4.56 & 202.13 & 5.44 & 5.70 & 6.35 & 5.14 & 23.66 & 38.25 \\ 
        & \checkmark &  &  & \fst{13.83} & 4.52 & \fst{202.02} & 5.39 & 5.63 & 6.33 & 5.12 & 23.66 & 38.22 \\ 
        & \checkmark & \checkmark &  & 15.33 & 9.02 & 207.61 & 6.45 & 6.09 & 7.56 & 5.64 & 24.08 & 39.82 \\ 
        & \checkmark & \checkmark & \checkmark & 13.85 & \fst{4.48} & \fst{202.02} & \fst{5.37} & \fst{5.53} & \fst{6.31} & \fst{5.11} & \fst{23.63} & \fst{38.12} \\
        \bottomrule
        \SetCell[c=9]{l}{\textbf{\textit{Hard level}}} \\
        \hline
        & & & & 21.62 & 30.06 & 253.94 & 11.69 & 7.38 & 7.07 & 7.01 & 30.50 & 43.54 \\ 
        \checkmark &  &  &  & 21.26 & 28.60 & 253.42 & \fst{11.46} & 7.25 & 7.12 & 6.95 & 29.95 & 43.03 \\ 
        & \checkmark &  &  & 21.25 & 28.55 & 253.17 & 11.56 & 7.25 & 7.10 & 6.91 & 29.92 & 43.01 \\ 
        & \checkmark & \checkmark &  & 24.97 & 45.62 & 260.08 & 14.62 & 8.55 & 9.92 & 8.17 & 30.66 & 48.03 \\ 
        & \checkmark & \checkmark & \checkmark & \fst{21.23} & \fst{28.49} & \fst{252.87} & 11.53 & \fst{7.24} & \fst{7.08} & \fst{6.89} & \fst{29.90} & \fst{42.98} \\
    \bottomrule
    \end{tblr}    
    \label{tab:sup_video_abl}
    \vspace{-1.5em}
\end{table}

\section{Video matting}

This section elaborates on the video matting aspect of our work, providing details about dataset generation and offering additional quantitative and qualitative analyses. For an enhanced viewing experience, we recommend visit our website, which contains video samples from V-HIM60 and real video results of our method compared to baseline approaches.

\subsection{Dataset generation}

To create our video matte dataset, we utilized the BG20K dataset for backgrounds and incorporated video backgrounds from VM108. We allocated 88 videos for training and 20 for testing, ensuring each video was limited to 30 frames. To maintain realism, each instance within a video displayed an equal number of randomly selected frames from the source videos, with their sizes adjusted to fit within the background height without excessive overlap.

We categorized the dataset into three levels of difficulty, based on the extent of instance overlap:
\begin{itemize}
    \item \textbf{Easy Level:} Features 2-3 distinct instances per video with no overlap.
    \item \textbf{Medium Level:} Includes up to 5 instances per video, with occlusion per frame ranging from 5 to 50\%.
    \item \textbf{Hard Level:} Also comprises up to 5 instances but with a higher occlusion range of 50 to 85\%, presenting more complex instance interactions.
\end{itemize}

During training, we applied dilation and erosion kernels to binarized alpha mattes to generate input masks. For testing purposes, masks were created using the XMem technique, based on the first-frame binarized alpha matte.

We have prepared examples from the testing dataset across all three difficulty levels, which can be viewed in the website for a more immersive experience. The datasets V-HIM2K5 and V-HIM60 will be made publicly available following the acceptance of this work.

\begin{table}[t]
    \centering
    \scriptsize
    \caption{\textbf{Our framework outperforms baselines in almost metrics on V-HIM60} (Extension of~\Tref{tab:video_matting}). We extend the result in the main paper with more metrics and our model is the best overall. \textbf{Bold} and \underline{underline} indicates the best and second-best model among baselines in the same test set.}
    \begin{tblr}{width=\columnwidth,colsep=2pt,colspec={@{}l|ccccccccc@{}}}
        \toprule
        Model &  MAD & MAD$_f$ & MAD$_u$ & MSE & SAD & Grad & Conn & dtSSD & MESSDdt \\
        \hline
        \SetCell[c=9]{l}{\textbf{\textit{Easy level}}} \\
        \hline
        MGM-TCVOM & 11.36 & \snd{18.49} & 202.28 & 5.13 & 4.11 & 4.57 & 3.83 & 17.02 & 19.69 \\ 
        MGM$^\star$-TCVOM & \snd{10.97} & 20.33 & \fst{187.59} & \snd{5.04} & \snd{3.98} & \snd{4.19} & \snd{3.70} & \snd{16.86} & \fst{15.63} \\ 
        InstMatt & 13.77 & 38.17 & 219.00 & 5.32 & 4.96 & 4.95 & 3.98 & 17.86 & 18.22 \\ 
        SparseMat & 12.02 & 21.00 & 205.41 & 6.31 & 4.37 & 4.49 & 4.11 & 19.86 & 24.75 \\ 
        Ours & \fst{10.12} & \fst{12.60} & \snd{192.63} & \fst{4.02} & \fst{3.68} & \fst{4.08} & \fst{3.43} & \fst{16.40} & \snd{16.41} \\ 
        \bottomrule
        \SetCell[c=9]{l}{\textbf{\textit{Medium level}}} \\
        \hline
        MGM-TCVOM & 14.76 & 4.92 & 218.18 & 5.85 & 5.86 & 7.17 & 5.41 & \fst{23.39} & \snd{39.22} \\ 
        MGM$^\star$-TCVOM & \fst{13.76} & \snd{4.61} & \fst{201.58} & \snd{5.50} & \fst{5.49} & \snd{6.47} & \fst{5.02} & 23.99 & 42.71 \\ 
        InstMatt & 19.34 & 35.05 & 223.39 & 7.50 & 7.55 & 7.21 & 6.02 & 24.98 & 54.27 \\ 
        SparseMat & 18.20 & 10.59 & 250.89 & 10.06 & 7.30 & 8.03 & 6.87 & 30.19 & 85.79 \\ 
        Ours & \snd{13.85} & \fst{4.48} & \snd{202.02} & \fst{5.37} & \snd{5.53} & \fst{6.31} & \snd{5.11} & \snd{23.63} & \fst{38.12} \\ 
        \bottomrule
        \SetCell[c=9]{l}{\textbf{\textit{Hard level}}} \\
        \hline
        MGM-TCVOM & \snd{22.16} & \snd{31.89} & 271.27 & \snd{11.80} & \snd{7.65} & 7.91 & \snd{7.27} & \snd{31.00} & 47.82 \\ 
        MGM$^\star$-TCVOM & 22.59 & 36.01 & \snd{264.31} & 13.03 & 7.75 & \snd{7.86} & 7.32 & 32.75 & \fst{37.83} \\ 
        InstMatt & 27.24 & 58.23 & 275.07 & 14.40 & 9.23 & 7.88 & 8.02 & 31.89 & 47.19 \\ 
        SparseMat & 24.83 & 32.26 & 312.22 & 15.87 & 8.53 & 8.47 & 8.19 & 36.92 & 55.98 \\ 
        Ours & \fst{21.23} & \fst{28.49} & \fst{252.87} & \fst{11.53} & \fst{7.24} & \fst{7.08} & \fst{6.89} & \fst{29.90} & \snd{42.98} \\
    \bottomrule
    \end{tblr}    
    \label{tab:sup_video_detail}
    \vspace{-1.5em}
\end{table}

\subsection{Training details}
For video dataset training (V-HIM2K5), we initialized our model with weights from the image pretraining phase. The training involved processing approximately 2.5M frames, using a batch size of 4 and a frame sequence length of $T=5$ on 8 A100 GPUs. We adjusted the learning rate to $5 \times 10^{-5}$, maintaining the cosine learning rate decay with a 1,000-iteration warm-up. In addition to the image augmentations, we incorporated motion blur (from OTVM) during training. Image sizes are kept the same as previously.
The first 3,000 iterations continued to use curriculum learning. In addition to the image augmentations, we incorporated motion blur (from OTVM) during training. For testing, the frame size was standardized to a short-side length of 576 pixels.

\subsection{Quantitative details}

Our ablation study, detailed in~\Tref{tab:sup_video_abl}, focuses on various temporal consistency components. The results demonstrate that our proposed combination of Bi-Conv-GRU and forward-backward fusion outperforms other configurations across all metrics. Additionally, \Tref{tab:sup_video_detail} compares our model's performance against previous baselines using various error metrics. Our model consistently achieves the lowest error rates in almost all metrics.

\begin{table*}[ht!]
    \centering
    \scriptsize
    \caption{\textbf{Our framework also reduces the errors of trimap propagation baselines.} When replacing those models' matte decoders with ours, the number in all error metrics was reduced by a large margin. Gray rows denote the module from public weights without retraining on our V-HIM2K5 dataset.}
    \begin{tblr}{width=\textwidth,row{3,6,10,13,17,20}={gainsboro},colspec={@{}cc|ccccccccc@{}}}
        \toprule
        Trimap prediction & Matte decoder &  MAD & MAD$_f$ & MAD$_u$ & MSE & SAD & Grad & Conn & dtSSD & MESSDdt \\
        \hline
        \SetCell[c=9]{l}{\textbf{\textit{Easy level}}} \\
        \hline
        OTVM & OTVM & 204.59 & 6.65 & 208.06 & 192.00 & 76.90 & 15.25 & 76.36 & 46.58 & 397.59 \\ 
        OTVM & OTVM & 36.56 & 299.66 & 382.45 & 29.08 & 14.16 & 6.62 & 14.01 & 24.86 & 69.26 \\ 
        OTVM & Ours & 31.00 & 260.25 & 326.53 & 24.58 & 12.15 & 5.76 & 11.94 & 22.43 & 55.19 \\
        \hline
        FTP-VM & FTP-VM  & 12.69 & 9.13 & 233.71 & 5.37 & 4.66 & 6.03 & 4.27 & 19.83 & 18.77 \\ 
        FTP-VM & FTP-VM & 13.69 & 24.54 & 269.88 & 6.12 & 5.07 & 6.69 & 4.78 & 20.51 & 22.54 \\ 
        FTP-VM & Ours & 9.03 & 4.77 & 194.14 & 3.07 & 3.31 & 3.94 & 3.08 & 16.41 & 15.01 \\ 
        \bottomrule
        \SetCell[c=9]{l}{\textbf{\textit{Medium level}}} \\
        \hline 
        OTVM & OTVM & 247.97 & 14.20 & 345.86 & 230.91 & 98.51 & 21.02 & 97.74 & 66.09 & 587.47 \\ 
        OTVM & OTVM & 48.59 & 275.62 & 416.63 & 37.29 & 17.25 & 10.19 & 17.03 & 36.06 & 80.38 \\ 
        OTVM & Ours & 36.84 & 209.77 & 333.61 & 27.52 & 13.04 & 8.63 & 12.69 & 32.95 & 70.84 \\ 
        \hline
        FTP-VM & FTP-VM  & 40.46 & 32.59 & 287.53 & 28.14 & 15.80 & 12.18 & 15.13 & 32.96 & 125.73 \\ 
        FTP-VM & FTP-VM & 26.86 & 28.73 & 318.43 & 15.57 & 10.52 & 12.39 & 9.95 & 32.64 & 126.14 \\ 
        FTP-VM & Ours & 18.34 & 11.02 & 234.39 & 9.39 & 6.97 & 6.83 & 6.59 & 26.39 & 50.31 \\ 
        \bottomrule
        \SetCell[c=9]{l}{\textbf{\textit{Hard level}}} \\
        \hline
        OTVM & OTVM & 412.41 & 231.38 & 777.06 & 389.68 & 146.76 & 29.97 & 146.11 & 90.15 & 764.36 \\ 
        OTVM & OTVM & 140.96 & 1243.20 & 903.79 & 126.29 & 47.98 & 17.60 & 47.84 & 59.66 & 298.46 \\ 
        OTVM & Ours & 123.01 & 1083.71 & 746.38 & 111.16 & 41.52 & 16.41 & 41.24 & 55.78 & 257.28 \\
        \hline
        FTP-VM & FTP-VM  & 46.77 & 66.52 & 399.55 & 33.72 & 16.33 & 14.40 & 15.82 & 45.04 & 76.48 \\ 
        FTP-VM & FTP-VM & 48.11 & 95.17 & 459.16 & 35.56 & 16.51 & 14.87 & 16.12 & 45.29 & 78.66 \\ 
        FTP-VM & Ours & 30.12 & 62.55 & 326.61 & 19.13 & 10.37 & 8.61 & 10.07 & 36.81 & 66.49 \\ 
        \bottomrule
    \end{tblr}    
    \label{tab:sup_video_trimap}
    \vspace{-1.5em}
\end{table*}

\begin{figure*}[t]
    \centering
    \includegraphics[width=\textwidth]{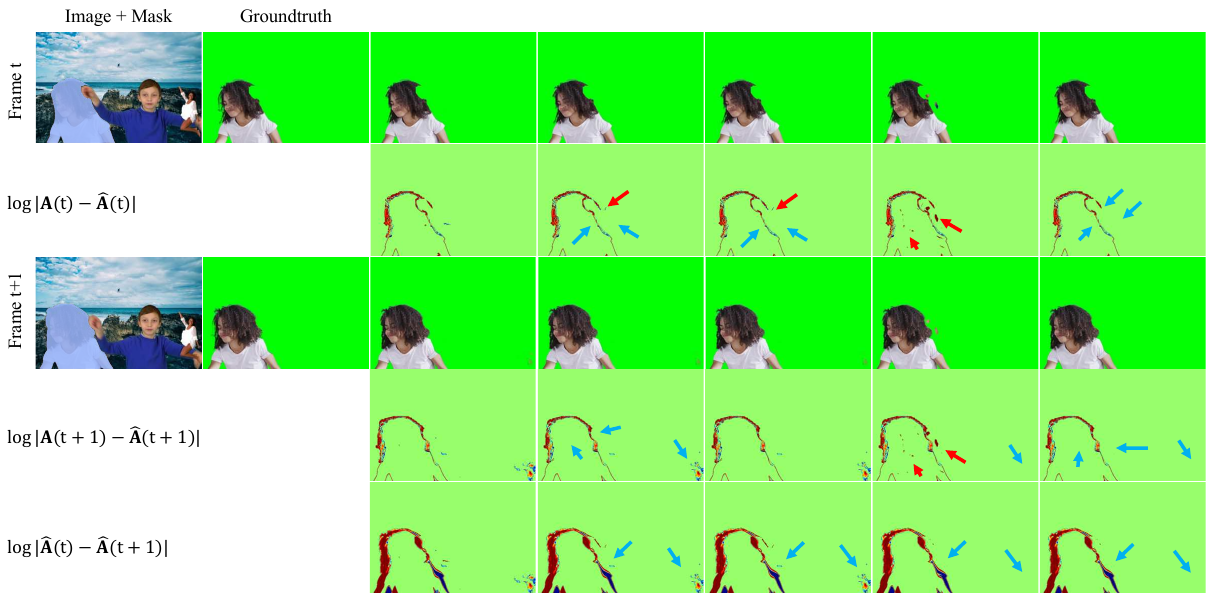}
    \begin{tblr}{width=\textwidth,colspec={X[1.2,c]|X[1,c]|X[1,c]X[1,c]X[1,c]X[1,c]X[1,c]}}
        \toprule
        \SetCell[r=2]{c}{Conv-GRU} & Single & & \checkmark \\
        \hline
        & Bidirectional & & & \checkmark & \checkmark & \checkmark \\
        \bottomrule
        \SetCell[r=2]{c}{Fusion} & $\mathbf{\hat{A}}^f$ & & & & \checkmark & \checkmark \\
        \hline
        & $\mathbf{\hat{A}}^b$ & & & & & \checkmark \\
        \bottomrule
    \end{tblr}
    \caption{\textbf{The effectiveness of different temporal components on the medium level of V-HIM60.}  Conv-GRU can improve the result a bit, but not perfect. Our proposed fusion strategy improves the output in both foreground and background regions. The table below denotes temporal components for each column. {\color{red} Red}, {\color{myblue} blue} arrows indicate the errors and improvements in comparison with the result without any temporal module. We also visualize the error to the groundtruth ($\log|\mathbf{A} - \mathbf{\hat{A}}|$) and the difference between consecutive predictions($\log|\mathbf{\hat{A}} - \mathbf{\hat{A}}|$). The color-coded map (min-max range) to illustrate differences between consecutive frames is \includegraphics[height=8pt]{fig/color_map.png}. (Best viewed in color and digital zoom).}
    \label{fig:sup_video_abl_compare}
\end{figure*}

An illustrative comparison of the impact of different temporal modules is presented in~\Fref{fig:sup_video_abl_compare}. The addition of Conv-GRU significantly reduces noise, although some residual noise remains. Implementing forward fusion $\mathbf{\hat{A}}^f$ enhances temporal consistency but also propagates errors from previous frames. This issue is effectively addressed by integrating $\mathbf{\hat{A}}^b$, which balances and corrects these errors, improving overall performance.

In an additional experiment, we evaluated trimap-propagation matting models (OTVM~\cite{seong2022otvm}, FTP-VM~\cite{huang2023ftpvm}), which typically receive a trimap for the first frame and propagate it through the remaining frames. To make a fair comparison with our approach, which utilizes instance masks for each frame, we integrated our model with these trimap-propagation models. The trimap predictions were binarized and used as input for our model. The results, as shown in~\Tref{tab:sup_video_trimap}, indicate a significant improvement in accuracy when our model is used, compared to the original matte decoder of the trimap-propagation models. This experiment underscores the flexibility and robustness of our proposed framework, which is capable of handling various mask qualities and mask generation methods.

\subsection{More qualitative results}

\begin{figure*}[ht!]
    \centering
    \vspace{-1em}
    \footnotesize
    \includegraphics[width=\textwidth]{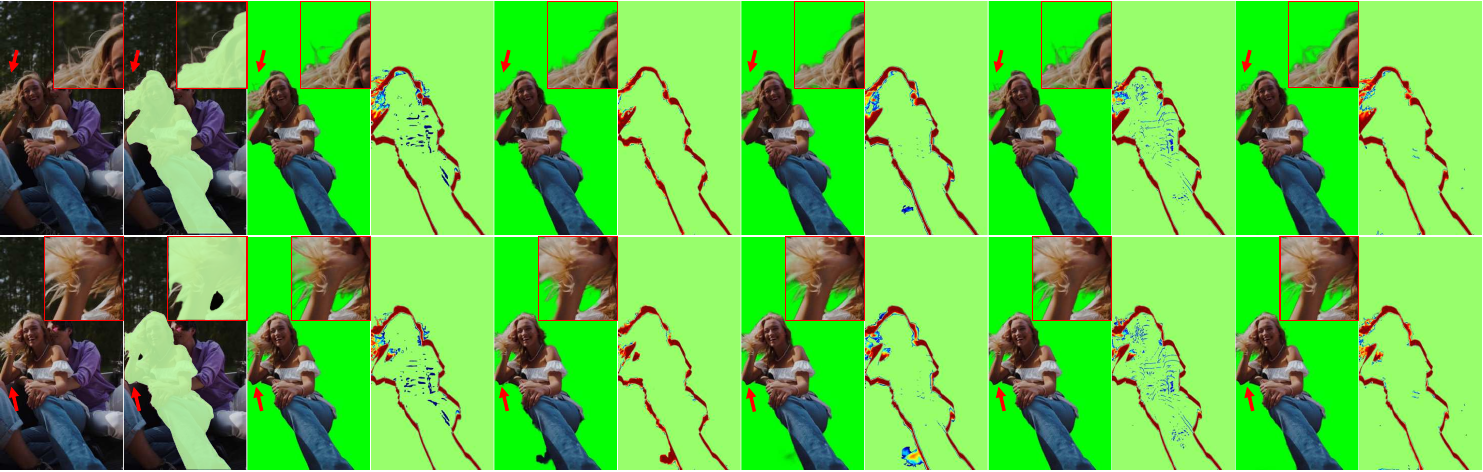}
    \includegraphics[width=\textwidth]{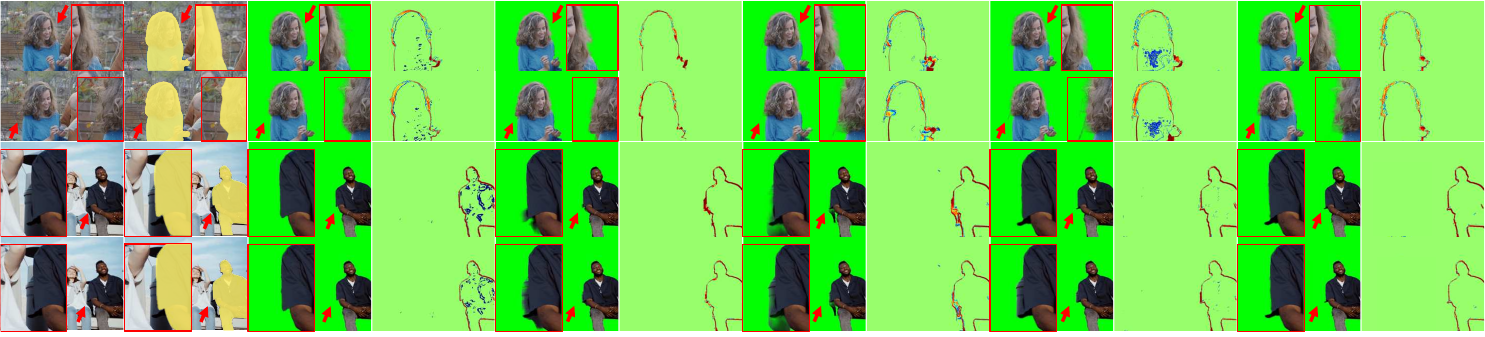}
    \begin{tblr}{width=\textwidth,rowsep=-6pt,colsep=3pt,colspec={X[1,c]X[1,c]X[1,c]X[1,c]X[1,c]X[1,c]X[1,c]X[1,c]X[1,c]X[1,c]X[1,c]X[1,c]X[1,c]}}
        \SetCell[r=2]{c}{Frame} & \SetCell[r=2]{c}{Input mask} & 
            $\hat{\mathbf{A}}$ & $\log\lft|\Delta_{\hat{\mathbf{A}}}\rgt|$ & 
            $\hat{\mathbf{A}}$ & $\log\lft|\Delta_{\hat{\mathbf{A}}}\rgt|$ & 
            $\hat{\mathbf{A}}$ & $\log\lft|\Delta_{\hat{\mathbf{A}}}\rgt|$ & 
            $\hat{\mathbf{A}}$ & $\log\lft|\Delta_{\hat{\mathbf{A}}}\rgt|$ & 
            $\hat{\mathbf{A}}$ & $\log\lft|\Delta_{\hat{\mathbf{A}}}\rgt|$ \\
        \vspace{-5cm}
        & & \SetCell[c=2]{c}{$\underbrace{\hspace{2.6cm}}_{\substack{\vspace{-5.0mm}}\colorbox{white}{~~InstMatt~~}}$}  &
        & \SetCell[c=2]{c}{$\underbrace{\hspace{2.6cm}}_{\substack{\vspace{-5.0mm}}\colorbox{white}{~~SparseMat~~}}$} &
        & \SetCell[c=2]{c}{$\underbrace{\hspace{2.6cm}}_{\substack{\vspace{-5.0mm}}\colorbox{white}{~~MGM-TCVOM~~}}$} &
        & \SetCell[c=2]{c}{$\underbrace{\hspace{2.6cm}}_{\substack{\vspace{-5.0mm}}\colorbox{white}{~~MGM$^\star$-TCVOM~~}}$} &
        & \SetCell[c=2]{c}{$\underbrace{\hspace{2.6cm}}_{\substack{\vspace{-5.0mm}}\colorbox{white}{~~Ours~~}}$} \\
    \end{tblr}
    \caption{\textbf{Highlighted detail and consistency on natural video outputs}. To watch the full videos, please check our website. We present the foreground extracted and the difference to the previous frame output for each model. The color-coded map (min-max range) to illustrate differences between consecutive frames is \includegraphics[height=8pt]{fig/color_map.png}.  {\color{red} Red} arrows indicate the zoom-in region in the {\color{red} red} square. (Best viewed in color and digital zoom).}
    \label{fig:sup_video_qual}
    \vspace{-1.5em}
\end{figure*}

For a more immersive and detailed understanding of our model's performance, we recommend viewing the examples on our website which includes comprehensive results and comparisons with previous methods. Additionally, we have highlighted outputs from specific frames in~\Fref{fig:sup_video_qual}.

Regarding temporal consistency, SparseMat and our framework exhibit comparable results, but our model demonstrates more accurate outcomes. Notably, our output maintains a level of detail on par with InstMatt, while ensuring consistent alpha values across the video, particularly in background and foreground regions. This balance between detail preservation and temporal consistency highlights the advanced capabilities of our model in handling the complexities of video instance matting.

For each example, the first-frame human masks are generated by r101\_fpn\_400e and propagated by XMem for the rest of the video.

\end{document}